\documentclass[11pt]{article}

\usepackage{multicol}
\usepackage{graphicx}
\usepackage{geometry}
\usepackage{stmaryrd}
\usepackage{amsfonts}
\usepackage{array}
\usepackage{amsmath}
\usepackage{bbm}
\usepackage{mathrsfs}
\usepackage[utf8]{inputenc}
\usepackage[english]{babel}
\makeatletter\AtBeginDocument{\let\@elt\relax}\makeatother
\usepackage[numbers]{natbib}
\usepackage{hyperref}
\usepackage{amsthm}

\hypersetup{
    colorlinks=true,
    citecolor=blue,
    linkcolor=red,
    pdfpagemode=FullScreen,
    }

\pdfpageattr {/Group << /S /Transparency /I true /CS /DeviceRGB>>}

\DeclareSymbolFont{matha}{OML}{txmi}{m}{it}
\DeclareMathSymbol{\varv}{\mathord}{matha}{118}

\makeatletter
\newcommand{\thickhline}{%
    \noalign {\ifnum 0=`}\fi \hrule height 1pt
    \futurelet \reserved@a \@xhline
}
\newcolumntype{"}{@{\hskip\tabcolsep\vrule width 1pt\hskip\tabcolsep}}
\makeatother

\makeatletter
\newcommand{\myfnsymbol}[1]{%
  \expandafter\@myfnsymbol\csname c@#1\endcsname
}

\newcommand{\@myfnsymbol}[1]{%
  \ifcase #1
  	
  \or a 
  \or b 
  \or c 
  \or d 
  \or \TextOrMath{\textasteriskcentered}{*}
  \fi
}

\newcommand{\affiliationTP}{\@myfnsymbol{1}}
\newcommand{\affiliationLMD}{\@myfnsymbol{2}}
\newcommand{\affiliationIECL}{\@myfnsymbol{3}}
\newcommand{\affiliationCMAP}{\@myfnsymbol{4}}
\newcommand{\correspondingAuthor}{\@myfnsymbol{5}}

\newcommand{\arlag}{l}
\newcommand{\predhor}{m}
\newcommand{\nwplag}{r}

\newcommand{\nobs}{n}

\title{Wind power predictions from nowcasts to 4-hour forecasts: a learning approach with variable selection}

\date{}

\author{
    Dimitri Bouche\textsuperscript{\affiliationTP \correspondingAuthor}, 
    Rémi Flamary\textsuperscript{\affiliationCMAP},
    Florence d'Alché-Buc\textsuperscript{\affiliationTP},
    Riwal Plougonven\textsuperscript{\affiliationLMD}, \\ 
    Marianne Clausel\textsuperscript{\affiliationIECL},
    Jordi Badosa\textsuperscript{\affiliationLMD},  
    Philippe Drobinski\textsuperscript{\affiliationLMD}}

\begin{document}

\renewcommand{\thefootnote}{\myfnsymbol{footnote}}
\maketitle

\footnotetext[1]{LTCI, Télécom Paris, Institut Polytechnique de Paris}%
\footnotetext[2]{LMD-IPSL, Ecole polytechnique - IP Paris, ENS - PSL Université, Sorbonne
Université, CNRS, France}%
\footnotetext[3]{Université de Lorraine, CNRS, IECL}%
\footnotetext[4]{CMAP, Ecole Polytechnique,  Institut Polytechnique de Paris}
\footnotetext[5]{\textbf{Corresponding author}: dimi.bouche@gmail.com}

\setcounter{footnote}{0}
\renewcommand{\thefootnote}{\fnsymbol{footnote}}

\abstract{
We study short-term prediction of wind speed and wind power (every 10 minutes up to 4 hours ahead). Accurate forecasts for these quantities are crucial to mitigate the negative effects of wind farms' intermittent production on energy systems and markets. We use machine learning to combine outputs from numerical weather prediction models with local observations. The former provide valuable information on higher scales dynamics while the latter gives the model fresher and location-specific data. So as to make the results usable for practitioners, we focus on well-known methods which can handle a high volume of data. We study first variable selection using both a linear technique and a nonlinear one. Then we exploit these results to forecast wind speed and wind power still with an emphasis on linear models versus nonlinear ones. For the wind power prediction, we also compare the indirect approach (wind speed predictions passed through a power curve) and the indirect one (directly predict wind power). 
}
\vspace{0.4cm}
\vspace{0.4cm}
\\
\textit{\textbf{Abbreviations---}}Numerical weather prediction (NWP); machine learning (ML); European Centre for Medium-Range Weather Forecasts (ECMWF); parc de Bonneval (BO); parc de Moulin de Pierre (MP); parc de Beaumont (BM); parc de la Renardière (RE); parc de la Vènerie (VE); kernel ridge regression (KRR); neural network (NN); ordinary least squares (OLS); forward stepwise ordinary least square (OLS f-stepwise); reproducing kernel Hilbert space (RKHS); Hilbert-Schmidt independence criterion (HSIC); Backward selection with HSIC (BAHSIC); normalized root mean squared error (NRMSE)

\section{Introduction}
The fast development of renewable energies is a necessity to mitigate climate changes \citep{AR6SP}. Wind energy has developed rapidly over the past three decades, with an average annual growth rate of 23.6\% between 1990 and 2016 \citep{IEA_Renewables2018}, and is now considered as a mature technology.  
The share of renewable energies in global electricity generation reached 29\% in 2020, and is expected to keep growing fast in coming years \citep{IEARenewables2021} which raises a number of challenges, stemming from the variability and spatial distribution of the resource. Then, in order to facilitate the dynamic management of electricity networks, forecasts of wind energy require continual improvement. Short timescales, from a few minutes to a few hours, are of particular importance for operations.

To produce forecasts, one can rely on several distinct sources of information. On timescales of half a day to about a week, deterministic weather forecasts provide a representation on a grid of the atmospheric state, including 
wind speed near the surface. The skill of such numerical weather forecasts (NWP) models has continually increased over the past decades \citep{BTB15}, while their spatial resolution has also grown finer (down to few km). However, to predict at a given geographical location for time horizons shorter than a day, the use NWP models is impeded by two main difficulties being \textbf{(i)} the errors in the modeled 
wind and \textbf{(ii)} the relatively infrequent initiation of forecasts. The former result from both limited resolution and the impossibility to model local processes. For instance, for wind speed at an altitude of 100m is strongly affected by local small-scale features and turbulent motions, both of which remain beyond the spatial resolution that is achievable for NWP models. Regarding the second point, operational centers typically launch forecasts twice or four times per day, however the computation of the forecast itself as well as the preparation of its initial state require time and computational resources--see e.g. \citep{Kalnay2003}. As a result, many methodologies for forecasting short-term wind speed or wind power use only past local observations and focus on statistical methods--see e.g. the reviews from \citet{TascikaraogluUzunoglu14, OkumusDinler2016, Liuetal2019} and references therein. Nevertheless, we know that NWP models can provide valuable information for the evolution of the atmosphere on larger scales--i.e. on the formation or passage of a low-pressure system and on the associated fronts.

It is therefore a natural idea to use both sources of information to train machine learning (ML) models: local deficiencies in NWP models can partly be overcome by {\it downscaling}; i.e. better estimating local variables from the knowledge from a model's outputs and past observations. Such efforts have been carried out for decades in meteorology and climatology, under different names. In a pioneering early study, \cite{GL72} applied multilinear regressions trained on past observations to correct NWP errors. More recently, \textit{model output statistics} has become common practice in operational weather forecast centers (see e.g. \cite{WilsonVallee2002,BaarsMass2005}). In recent years, ML methods have 
enhanced the performance of these post-processing steps (see e.g. \cite{ZBMS16, Gouthametal2021}).

{Specifically for wind speed or wind power forecasting, several \textit{hybrid models} combining successfully NWP outputs with local observations have been proposed. 
In terms of time horizons, the focus is mostly on forecasts beyond 1 hour with low resolution (generally one prediction per hour)--see e.g. \citet[Table 1]{HoolohanAl18gps} and references therein. While for the shorter term, most existing hybrid methods rely on complex and deep architectures--see e.g. \citet[Table 1]{HanAl22newdeep} and references therein. Moreover, for all these methods only a very low number of local and NWP variables are used (most of the times, only the past observed wind speeds and the ones predicted by the NWP model).

In this paper, we study hybrid prediction of both wind speed and wind power. Our contributions are five-fold.
\begin{itemize}
\item We study the problem for time horizons ranging from 10 minutes to 4 hours at a high resolution (every 10 minutes). This allows us to study with precision when and how the transition from one source of information (past local observations) to the other (NWP forecasts) occurs. This setting has been introduced in \citet{DupreAl20} yet we extend it and use it to address the following new problems.  
\item We include many different outputs from a NWP model as they could provide broader information on the overall predicted dynamics to the ML model. We also include several local variables. We then focus on variable selection and study the evolution of the importance of the selected variables through time. This allows us to better understand the nature of the studied statistical relationship and to extract a usable set of relevant variables.
\item We study five distinct wind farms which enables us to expose many similarities but also some site specificities and increases the statistical significance of our results. 
\item We investigate which type of ML methods are the most suited for hybrid prediction of wind speed and wind power. 
\item Many existing contribution focus either on wind power or wind speed prediction but not on the relation between them, whereas at all steps of the paper, we compare the direct (predict wind power) and indirect (wind speed predictions passed through a power curve) approaches.
\end{itemize}
In terms of methodology, we have two key focuses.
\begin{itemize}
\item We want this study to be usable by practitioners. To that end we concentrate on a reduced choice of well-known and efficient ML methods which scale well with the number of samples, and moreover provide all the needed elements for a straightforward implementation. We also put a particular emphasis on how we select our models. 
\item We want to ensure our results are statistically significant. To that end, we employ a thorough evaluation process. We study several sites over several periods of time (the number of samples is quite high per site) and for each location, we average the results over several data splits.
\end{itemize}}

In Section \ref{sec:context}, we introduce the data set from Zéphyr ENR and detail our processing of the data. Section \ref{sec:methodo} is dedicated to the presentation of our methodology as well as to the introduction of the statistical learning tools.
Then in Section \ref{sec:varsel} investigates variable selection. Finally in Section \ref{sec:forecasting}, exploiting all the previous results, we compare different well-known ML models as well as direct and indirect prediction for wind power.
\vspace{0.3cm}
\\
\noindent {\bf Notation} We introduce the following notation: for two integers $n_0, n_1 \in \mathbb N^*$, the set of strictly positive integers, we denote by $\llbracket n_0 \rrbracket $ the set $\{1,...,n_0 \}$ and by $\llbracket n_0, n_1 \rrbracket$ the set $\{n_0,...,n_1 \}$.

\section{Data and context} \label{sec:context}
In this Section, we introduce the dataset that we use (Section \ref{subsec:data}) as well as the pre-processing steps that we apply to it and the general evaluation methodology (Section \ref{subsec:data-methodo}).

\subsection{Zéphyr ENR's dataset} \label{subsec:data}
\begin{figure}
     \centering
     \includegraphics[width=0.7\textwidth]{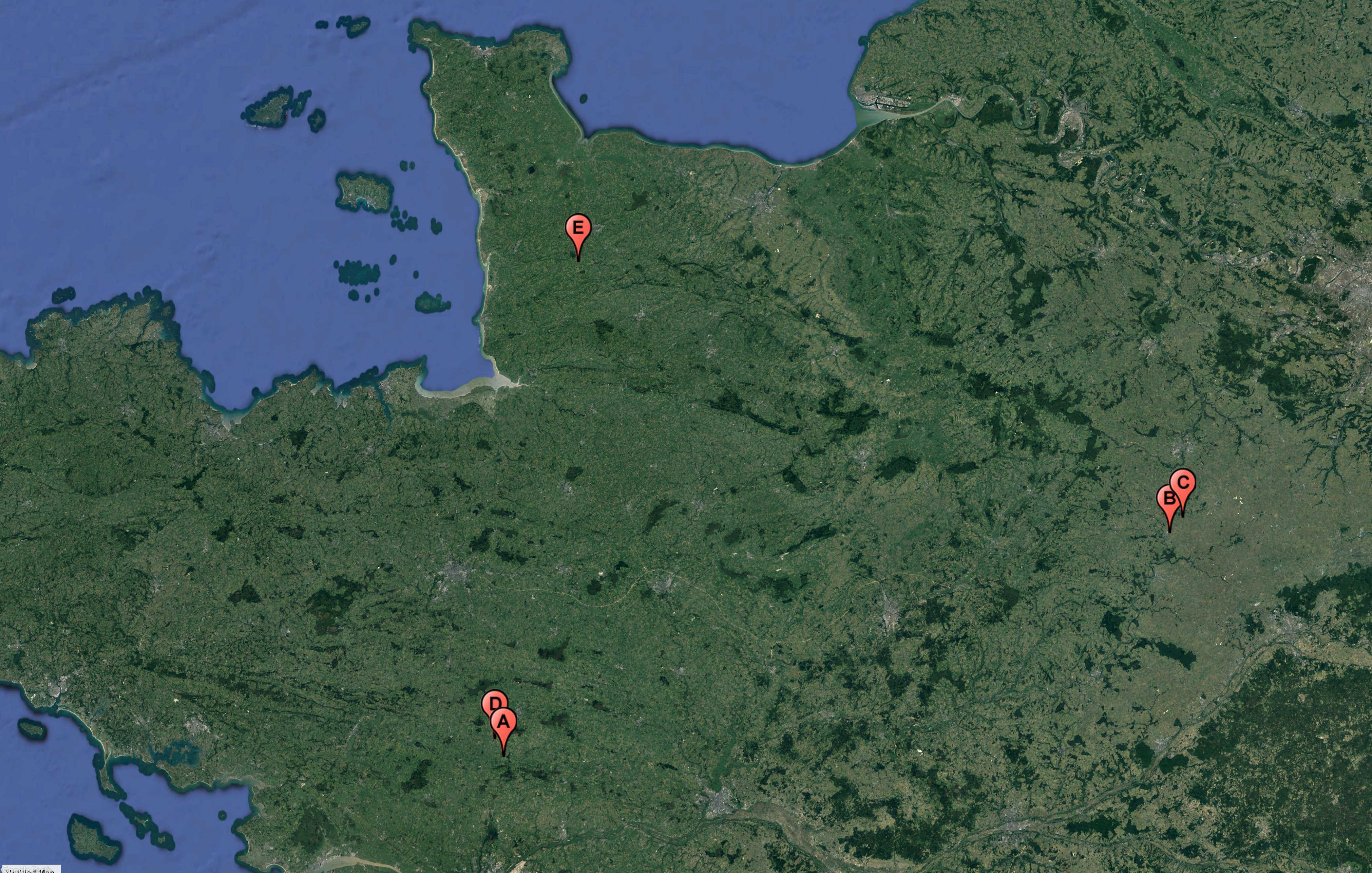}
     \caption{Cartography of the studied farms, BM (A), BO (B), MP (C), RE (D), VE (E)}
     \label{fig:sat}
\end{figure}

\begin{table}[h]
\centering
\begin{footnotesize}
	\begin{tabular}{llll}
		\thickhline
		Variable type & Altitude or pressure level & Variable & Unit \\
		\hline
		Surface & 10m/100m & Zonal wind speed & ms$^{-1}$ \\
		& & Meridional wind speed & ms$^{-1}$ \\
		& 2m & Temperature & K \\
		&  & Dew point temperature & K \\
		& Surface & Skin temperature & K \\
		& & Mean sea level pressure &  Pa \\
		& & Surface pressure & Pa \\
		& & Surface latent heat flux & Jm$^{-2}$ \\
		& & Surface sensible heat flux & Jm$^{-2}$ \\
		& & Boundary layer dissipation & Jm$^{-2}$ \\
		& & Boundary layer height & m \\
		\hline
		Altitude & 1000/925/850/700/500 & Zonal wind speed & ms$^{-1}$ \\
		& & Meridional wind speed & ms$^{-1}$ \\
		& & Geopotential height & m$^2$s$^{-2}$ \\
		& & Divergence & s$^{-1}$ \\
		& & Vorticity & s$^{-1}$ \\
		& & Temperature & K \\
		\hline
		Computed & 10m/100m & Norm of wind speed & ms$^{-1}$ \\
		& 10m to 925 hPa & Wind shear & ms$^{-1}$ \\
		& & Temperature gradient & K\\
	\end{tabular}
	\end{footnotesize}
    \caption{ECMWF variables}
    \label{tab:ecmwf}
\end{table}

\begin{table}[h]
\centering
\begin{footnotesize}
    \begin{tabular}{lll}
		\thickhline
		Availability & Variable & Unit \\
		\hline
		All & Wind speed & ms $^{-1}$ \\
		All & Power output & kW \\
		All & Wind direction & Degree \\
		BO and BM & Temperature & Celsius degree \\
	\end{tabular}
	\end{footnotesize}
    \caption{In situ variables}
    \label{tab:insitu}
\end{table}

\begin{table}[h]
\centering
\begin{footnotesize}
	\begin{tabular}{ll}
		\thickhline
		Variable (source) & Abbreviation \\
		\hline
		Wind speed (in situ) & WS \\
		Power output (in situ) & PW \\
		Norm of wind speed at 100m (ECMWF) & F10 \\
		Norm of wind speed at 100m (ECMWF) & F100 \\
		Wind shear between 10m and 925 hPa (ECMWF) & DF \\
		Boundary layer dissipation (ECMWF) & bld \\
		Boundary layer height (ECMWF) & blh \\
		Surface latent heat flux (ECMWF) & slhf \\
	\end{tabular}
	\end{footnotesize}
    \caption{Abbreviations for the variables used in the paper}
    \label{tab:abbreviations}
\end{table}

Our first source of information consists of measurements made by sensors in the wind turbines (we call these in situ variables) whereas the second one consists of forecasts from the European Centre for Medium-Range Weather Forecasts (ECMWF).
We study five wind farms in the northern half of France: Parc de Bonneval (BO), Moulin de Pierre (MP), Parc de Beaumont (BM), Parc de
la Renardière (RE), and Parc de la Vènerie (VE). These wind farms are operated
by the private company Z\'ephyr ENR and are described in details in
\citep{dupre2020economic}. We display their location on a map in Figure
\ref{fig:sat}. Some are geographically close by--we can form the pairs (BO, MP) and (BM, RE)--while VE is isolated. Note that we left another available farm out of the study because it displayed signs of sensors deficiencies. On the one hand, the geographical topology of the surroundings for (BO, MP) and (BM, RE) are quite similar, they correspond to open fields with very few
elevation variations. On the other hand, VE is surrounded by wooded
countryside with slightly more elevation variations, which may explain the
differences that we observe between this farm and the others in Sections
\ref{sec:varsel} and \ref{sec:forecasting}.

For BO and VE we have three years of data (from 2015 to 2017) which amounts to a total of $\nobs=157680$ observations for BO. However, for VE we do not use the year 2016 because it encompasses sensor deficiencies, so we use $\nobs=105120$ observations. For BM and RE we have access to two years of data (from 2017 to 2018 for BM and from 2015 to 2016 for RE) which results in a total of $\nobs=105120$ observations, and finally for MP we have only one year (2017), which gives us of total of $\nobs=52560$ observations. 

Several variables are available, we summarize the in situ ones in Table \ref{tab:insitu}--note the temperature is available only for BO and BM. In order to encode the circular nature of the in situ wind direction we encode it using two trigonometric variables.

The ECMWF provides global forecasts issued by their NWP models. We followed \citep{DupreAl20}: we extracted the day ahead forecast twice a day (at 0000UTC and 1200UTC) and included the same 47 variables as they do. These variables are either selected  or computed so as to describe as well as possible the boundary layer, the wind parameters and the temperature in the lower troposphere. Table \ref{tab:ecmwf} presents the ECMWF variables we use. They can be either surface variables, altitude ones, or computed from other ECMWF variables. The spatial resolution of ECMWF forecasts is about 16 km (0.125 $^\circ$ in latitude and longitude) and their temporal resolution is 1h, then to match that of the in situ variables (10 min), we linearly interpolate the ECMWF forecasts. To finish with we sum up the abbreviations for the variables mostly used in the paper in Table \ref{tab:abbreviations}. 

\subsection{Preprocessing and evaluation methodology} \label{subsec:data-methodo}

In order to increase the statistical significance of our results, we average the outcomes of the experiments over different data splits. A split consists of 3 subsets from the dataset, a train subset (of size $\nobs_{\text{train}}$), a validation one (size $\nobs_{\text{val}}$) and a test one (of size $\nobs_{\text{test}}$). {In order to avoid overfitting, given a ML method and a set of possible parameter values, we first train the resulting models on the train set. Then we choose the model yielding the best score on the validation set. To finish with, we re-train this model on the concatenation of the train and validation set and report its score on the test set.} So as to preserve time coherence, we build our splits in a rolling fashion. For instance for the first split we take the period $\llbracket \nobs_{\text{train}} \rrbracket$ for training, the period $\llbracket \nobs_{\text{train}} + 1, \nobs_{\text{train}} +\nobs_{\text{val}} \rrbracket$ for validation and we test the models on the period $\llbracket \nobs_{\text{train}} + \nobs_{\text{val}} + 1, \nobs_{\text{train}} +\nobs_{\text{val}} + \nobs_{\text{test}} \rrbracket$. Then for the second split, the train period is $\llbracket \nobs_{\text{train}} +\nobs_{\text{val}} + \nobs_{\text{test}} + 1, 2\nobs_{\text{train}} + \nobs_{\text{val}} + \nobs_{\text{test}} \rrbracket$, the validation one is $ \llbracket 2 \nobs_{\text{train}} +\nobs_{\text{val}} + \nobs_{\text{test}} + 1, 2\nobs_{\text{train}} + 2 \nobs_{\text{val}} + \nobs_{\text{test}} \rrbracket$ and so on. For the sizes of the windows, we set $\nobs_{\text{train}} = 10000$, $\nobs_{\text{val}} = 10000$ and
$\nobs_{\text{test}} = 10000$ (however, the last split generally
contains around $5000 \leq \nobs_{\text{test}} \leq 10000$ observations). Since the length of available data vary from farm to farm, we do not have the same number of splits for all the farms. 

We pre-process the data in the following way. As the number of wind turbines per farm is quite low (6 for BM, 6 for BO, 3 for HC, 6 for MP, 6 for RE and 4 for VE) 
, we average the in situ data over the turbines for each farm. 
In all our experiments, we standardize both the input and the output variables (subtract the mean and divide by the standard deviation) using the training data. We do so both for in situ variables and ECMWF ones. Such operation is crucial for instance to avoid favoring some variables which are structurally bigger over others when using regularized ML models.  

\section{Methodology and machine learning tools} \label{sec:methodo}
In this Section, we introduce our general methodology as well as the ML tools that we use for variable selection and forecasting.

\subsection{Methodology} \label{subsec:methodo}

\begin{figure}[h]
     \centering
     \includegraphics[width=0.9\textwidth]{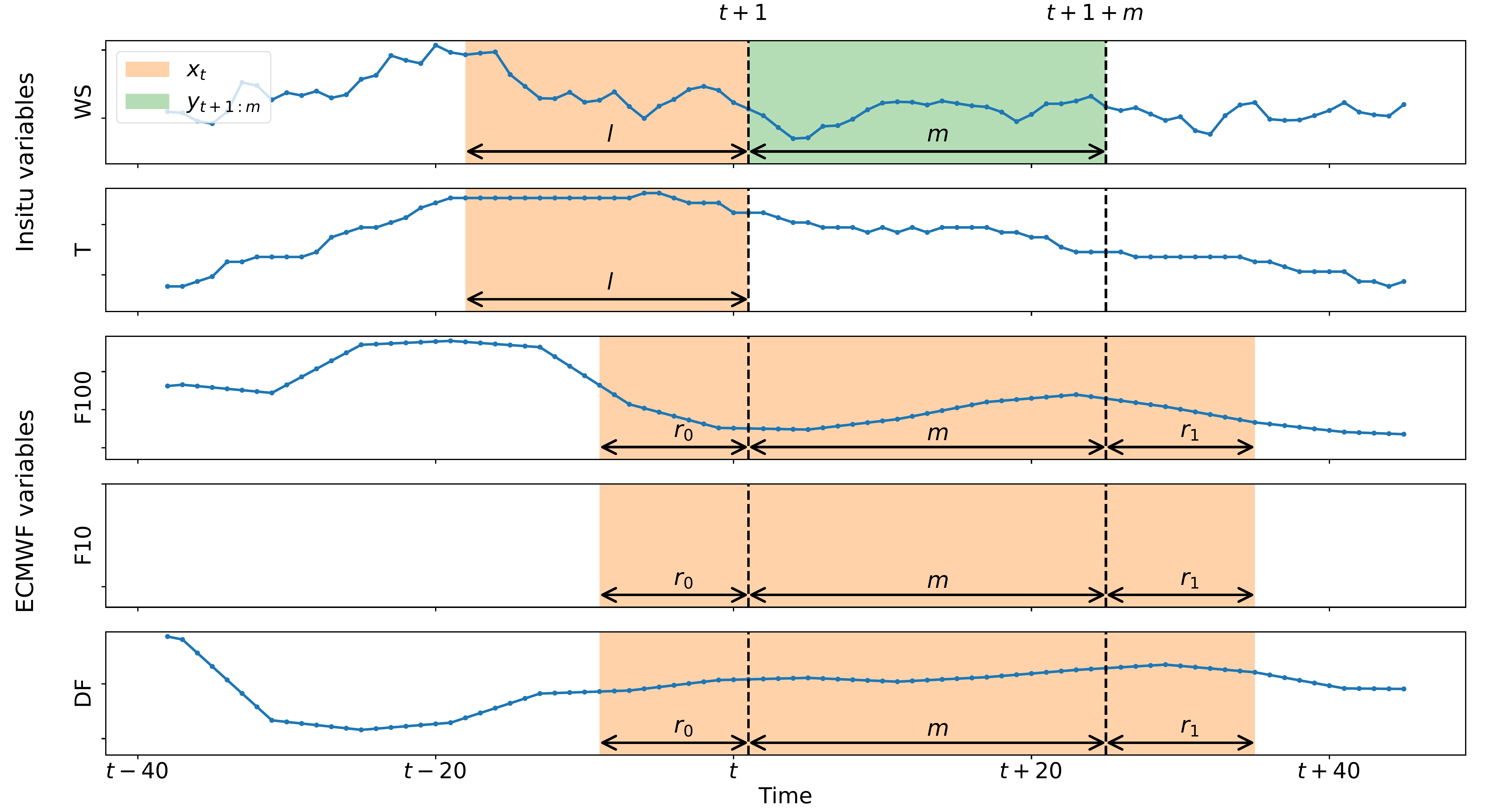}
     \parbox{\dimexpr\columnwidth-2.3cm}{
     \caption{Summary of the time windows used for each source of data for wind speed prediction}}
     \label{fig:windows}
\end{figure}

\noindent{\bf Dataset building.} Let $\predhor \in \mathbb N$ be the prediction length (the number of future wind speed or wind power values we want to forecast). For the ECMWF variables, we include the corresponding forecasts. However, in practice we found that including a bit more than that improved performances. To that end, we denote respectively by $\nwplag_0 \in \mathbb N$ the number of past ECMWF predictions that we include, and by $\nwplag_1 \in \mathbb N$ the number of ECMWF predictions that we consider after $\predhor$. For the in situ variables, we include a length $\arlag \in \mathbb N$ of past observations. These different time windows are illustrated in Figure \ref{fig:windows} for a reduced set of variables. For the all the variables and time windows, we concatenate the relevant observations $\mathbf x_t$ (these within the orange zones in Figure \ref{fig:windows}). From these, we want to produce a prediction for $\mathbf y_{t+1:m} \in \mathbb R^m$ (the $m$ observations within the green zone in Figure \ref{fig:windows}). In practice, we use the following parameters which work well experimentally:
\begin{itemize}
\item we predict up to 4 hours, with a time sampling rate of 10 min, it means that $\predhor= \frac {4 \times 60} {10} = 24$, 
\item for the in situ variables, we consider 3h of past observations, thus $\arlag = \frac {3 \times 60}{10} = 18$,
\item for the ECMWF variables we additionnally use the predictions between 1.5h before and 1.5h after the time horizon of interest so 
$\nwplag_0 = \nwplag_1 = \frac{1.5 \times 60}{10}$.
\end{itemize}

\noindent{\bf ML models.} We stick to ML models which are well-known and can scale well to a higher volume of data. Good results were obtained for one location (BO) from the studied dataset in \citet{DupreAl20} using linear regression with greedy forward stepwise variable selection \citep{HastieAl01}.
Nevertheless, since we are interested in the importance of variables, we study also an alternative which select variables directly in the least square problem. The LASSO \citep{Tibshirani96} exploits the L1 penalty to induce sparsity in the regression coefficients, thus shrinking to zero the ones which are associated to the less relevant variables. Such methods can however be limited in that they can learn only linear dependencies between the input and output variables. Consequently, we study a nonlinear alternative: kernel ridge regression (KRR)--see for instance \citep{ScholpkopfSmola02, Shawe-TaylorCristianini04}. Finally, in order to include most families of ML models, we consider two other nonlinear methods: a tree-based boosting algorithm \citep{Friedman01} (we use XG-Boost \citet{ChenGuestrin16}) as well as a feed-forward neural network (NN).
In Section \ref{subsec:ml-details} we give more mathematical details on the general ML problem as well as on the methods that perform the best in the experimental section.
\vspace{0.17cm}
\\
\noindent{\bf Variable selection.} We have many in situ and ECMWF variables available (see Tables \ref{tab:ecmwf} and \ref{tab:insitu}). So as to improve the computational efficiency and 
and understand better what the models do, it is preferable to use only the most important variables. Ideally, we want to find a subset which is sufficient for a model to predict a statistically relevant target value from the input variables. In that sense a variable selection tool is necessarily model specific. Linear techniques will focus only on linear dependencies, whereas nonlinear ones will incorporate a much wider range of dependencies. 
Then we propose to use and interpret the results of one variable selection for each type. For the linear one we study the LASSO. For the nonlinear one, we use backward elimination using the Hilbert Schmidt Independence Criterion \citep{SongAl12}. As opposed to the LASSO, it performs variable selection as an independent first step. The selected variables can then be used downstream to train any model. Then, for the nonlinear models (KRR, XG-Boost, feed-forward NN), we use the variables selected through this method. We give more detailed insights into the different techniques in Section \ref{subsec:varsel-details}.

\subsection{Details on machine learning models} \label{subsec:ml-details}

The input observations are the concatenation of the different variables on the time windows described in the previous section. We denote by $\mathcal X = \mathbb R^q$ the resulting input space, for some $q \in \mathbb N$. 
Our training data then consist of $((\mathbf x_t, \mathbf y_{t+1:m}))_{t=1}^n \in (\mathcal X \times \mathbb R^m)^{n}$ for some $n \in \mathbb N$, where we recall that $\mathbf y_{t+1:m} = (y_{t+1+m_0})_{m_0=1}^m$.
Given a prediction function from a ML model class $f_{\mathbf w}: \mathcal X \rightarrow \mathcal Y$ parameterized by a vector $\mathbf w \in \mathcal W$, we want to minimize the average error on the training data:

\begin{equation} \label{eq:rermp}
\min_{\mathbf w \in \mathcal W} \frac 1 {n} \sum_{t=1}^{n} \| f_{\mathbf w}(\mathbf x_t) - \mathbf y_{t+1:m} \|_2^2 + \lambda \Omega(\mathbf w).
\end{equation}

\noindent However, depending on the model, a penalty function $\Omega: \mathcal W \longrightarrow \mathbb R$ can be added in order to prevent overfitting or promote variable selection; its intensity is controlled by a parameter $\lambda > 0$.

In practice, instead of predicting all time horizons in one go as in Problem \eqref{eq:rermp}, we rather use separate models for each horizon in $\llbracket t+1, t+1+\predhor \rrbracket$. 
That way we can tailor the different parameters for each horizon, which we found improved performances. Then in what follows, we consider a generic time horizon $m$ and predict $y_{t+1+m}$.
\vspace{0.17cm}

\noindent{\bf Ordinary least squares (OLS).}
In forward stepwise variable selection \citep{HastieAl01}, at each step an OLS regression is solved for which the optimization problem reads:

\begin{equation} \label{eq:ols}
\min_{\mathbf w \in \mathcal W, b \in \mathbb R} \frac 1 n \sum_{t=1}^{n} (\mathbf w^{\text T} \mathbf x_t + b - y_{t+1+m})^2
\end{equation}

\noindent A well-known and simple closed form exist, which we use in practice. 
\vspace{0.17cm}

\noindent{\bf LASSO}. The optimization problem for the LASSO is the following:

\begin{equation*}
\min_{\mathbf w \in \mathcal W, b \in \mathbb R} \frac 1 n \sum_{t=1}^{n} ( \mathbf w^{\text T} \mathbf x_t + b - y_{t+1+m} )^2 + \lambda \| \mathbf w \|_1,
\end{equation*}

\noindent where $\| \mathbf w \|_1$ is the sum of the absolute values of the coefficients $\mathbf w$. Many efficient algorithms exist to solve this convex yet non differentiable problem--see for instance \citep{BeckTeboulle09}. In practice we use the scikit-learn \citep{Scikit11} implementation with coordinate descent solver. 
\vspace{0.17cm}

\noindent{\bf Kernel ridge regression (KRR).} In KRR, we consider a class of models defined by a positive definite reproducing kernel $k: \mathcal X \times \mathcal X \longrightarrow \mathbb R$ which results in a unique associated reproducing kernel Hilbert space (RKHS). A most typical choice for $k$ is the Gaussian kernel:

$$k_{\gamma}(\mathbf x, \mathbf x') := \exp \left ( - \gamma (\| \mathbf x - \mathbf x' \|_2^2 \right ).$$

\noindent We then seek our modeling function in this RKHS which we denote $\mathcal H_k$, each $h \in \mathcal H_k$ being a function from $\mathcal X$ to $\mathbb R$. For many kernels, this space constitutes a very rich class of modeling functions which can model nonlinear dependencies. The optimization problem reads:

\begin{equation} \label{eq:krr}
\min_{h \in \mathcal H_k} \frac 1 n \sum_{t=1}^{n} (h(\mathbf x_t) - y_{t+1+m} )^2 + \lambda \| h \|_{\mathcal H_k}^2
\end{equation}

\noindent where $\| \cdot \|_{\mathcal H_k}^2$ is the RKHS norm on $\mathcal H_k$, which measure in a sense the smoothness of functions in $\mathcal H_k$.
Thanks to the Representer theorem, any solution to Problem \eqref{eq:krr} can be parameterized by a vector $\pmb \alpha \in \mathbb R^{n}$:

\begin{equation*}
h_{\pmb \alpha} := \sum_{j=1}^{\nobs} \alpha_{j} k(\mathbf x_j, \cdot), 
\end{equation*}

\noindent which makes optimization in the RKHS amenable. For KRR the optimal coefficients $\widehat{\pmb \alpha}$ can be found in close form:

$$ \widehat{\pmb \alpha} := (K + n \lambda I)^{-1} \mathbf y^{(m)}, $$ 

\noindent with $\mathbf y^{(m)}:= (y_{t+1+m})_{t=1}^n$, $I \in \mathbb R^{n \times n}$ the identity matrix and $K \in \mathbb R^{n \times n}$ with entries $K_{tj}=k(\mathbf x_t, \mathbf x_j)$.

In practice, to handle the large volume of training data, we use an approximated version of KRR. Nyström approximation \citep{WilliamsSeeger01, DrineasMahoney05} exploits a random subset of points from the training data. Concretely, we sample randomly and uniformly without replacement $p \in \mathbb N$ indices $\{i_1,...,i_p \}$ among the integers in $\llbracket n \rrbracket$, and replace $h_{\widehat{\pmb \alpha}}$--see e.g. \citep{RudiAl15}--with:

$$ \widetilde h_{\widetilde{\pmb \alpha}}:= \sum_{j=1}^p \widetilde \alpha_j k(\mathbf x_{i_j}, \cdot), $$

\noindent where $\widetilde{\pmb \alpha} \in \mathbb R^p$ is given by the following close form:

\begin{equation} \label{eq:nystrom-krr}
\widetilde {\pmb \alpha}:= (K_{np}^{\text T} K_{np} + \lambda n K_{pp})^{\dagger} K_{np}^{\text T} \mathbf y^{(m)}.
\end{equation}

\noindent Where $A^\dagger$ denotes the Moore-Penrose pseudo-inverse of a matrix $A$, and $K_{np} \in \mathbb R^{n \times p}$ is defined by the entries $(K_{np})_{tj}:= k(\mathbf x_t, \mathbf x_{i_j})$ and $K_{pp} \in \mathbb R^{p \times p}$ by the entries $(K_{pp})_{bj} = k(\mathbf x_{i_b}, \mathbf x_{i_j})$.

\subsection{Details on variable selection} \label{subsec:varsel-details}

\noindent{\bf OLS with forward stepwise selection (OLS f-stepwise).} When performing linear regression, variable selection can be performed in a greedy manner. First an intercept is fit to the data and then at each step we solve OLS problems--Problem \eqref{eq:ols}--adding in turns each one of the remaining variables. We then keep the one which best improve the model according to some criterion. In \citep{DupreAl20}, the Bayesian information criterion is used. However, in our experiments we rather used the improvement of the score on half of the validation set, as it led to better experimental performances.
\vspace{0.17cm}

\noindent{\bf LASSO.} Provided the regularization intensity $\lambda$ is well chosen, the L1 penalty of the LASSO shrinks the coefficients associated the variables which are less important towards zero. Then the model uses mostly the relevant variables and the magnitude of the coefficients can be looked at to deduce what these variables are. This is the type of analysis that we perform in Section \ref{subsec:lasso-sel}.
\vspace{0.17cm}

\noindent{\bf Hilbert-Schmidt independence criterion (HSIC).} The HSIC \citep{GrettonHsic05} is an independence measure. Similarly to the KRR, it makes use of RKHSs to embed implicitly a set of
observations into a high-dimensional space and consider a notion of independence in this space which allows for detection of nonlinear dependencies.
More precisely, let us consider a positive-definite kernel
$k: \mathcal X^2 \longrightarrow \mathbb R$ for the input observations and a one $g: (\mathbb R^m )^2 \longrightarrow \mathbb R$ for the output observations. For this variable selection technique, we consider all time horizons in $\llbracket t+1,t+1+\predhor \rrbracket$ together as the kernelized framework allows for this. In practice, we estimate HSIC from the data as \citep{GrettonAl08}:

\begin{equation*}
\widehat{\text{HSIC}}:= \frac 1 {n^2} \text{Trace}(H K H G), 
\end{equation*}

\noindent where $H \in \mathbb R^{n \times n}$ is the centering matrix  $H: = \frac 1 n (I - \pmb 1 \pmb 1^{\text T})$ with $\pmb 1 \in \mathbb R^{n}$ a vector full of ones and $I \in \mathbb R^{n \times n}$ the identity matrix. The matrices $K \in \mathbb R^{n \times n}$ and $G \in \mathbb R^{n \times n}$ are the kernel matrices:

\begin{equation*}
\begin{split}
(K)_{tj} & := k (\mathbf x_t, \mathbf x_j), \\
(G)_{tj} & := g (\mathbf y_{t+1:m}, \mathbf y_{j+1:m}).
\end{split}
\end{equation*}

\noindent HSIC takes its values between $0$ and $1$, a value of $0$ meaning independence and a value of $1$ means full dependence. 

However, to be able to compute the estimator for the large number of data points, we recourse to Nyström approximation as well \citep{ZhangAl18}. We then sample randomly and uniformly without replacement $p \in \mathbb N$ indices $\{i_1,...,i_p\}$ from the integers in $\llbracket n \rrbracket$ for the input observations and $p' \in \mathbb N$ ones $\{i'_1,...,i'_{p'} \}$ for the output observations. We then define the Nyström features maps \citep{TianbaoAl12nystrom} (centered in the feature space using $H$):

\begin{equation*}
\begin{split}
\widehat{\Phi} & := H K_{np} K_{pp}^{-\frac 1 2}, \\
\widehat {\Psi} & := H G_{np'} G_{p'p'}^{-\frac 1 2}, 
\end{split}
\end{equation*}

\noindent where the matrices $K_{np}$ and $K_{pp}$ are defined as for Equation \eqref{eq:nystrom-krr} and the matrices $G_{np'}$ and $G_{p'p'}$ are defined similarly for the kernel $g$ however based on the set of indices $\{i'_1,...,i'_{p'} \}$. The Nystr\"om HSIC estimator is then \citep{ZhangAl18}:

$$ \widetilde{\text{HSIC}}:= \| \frac 1 n \widehat \Phi^{\text T} \widehat \Psi \|_F^2,$$

\noindent where for a matrix $A$, the Frobenius norm is defined as $\| A \|^2_F:= \text{Trace}(A^{\text T} A)$.
\vspace{0.17cm}
\\
\noindent{\bf Backward selection with HSIC (BAHSIC).} To perform variable selection, we start with all the available variables and then at each round, we compute the
HSICs between the input variables and the target variable removing one input variable at a time. A given percentage of the input variables for which these HSICs are the highest are removed. We keep iterating in this way to rank the variables. Then, the ones removed the latest are the most important ones. The detailed algorithm corresponds to Algorithm 1 in \citep{SongAl12}. A forward version exists as well, however, the authors advocate the use of backward selection to avoid missing important variables. Finally as a side note, in practice we use as Gaussian kernels setting bandwidth following the recommendations from \citep{SongAl12}.

\section{Importance of variables and their evolution through time} \label{sec:varsel}
In this section, we study variable selection using the LASSO in Section \ref{subsec:lasso-sel} and BAHSIC in Section \ref{subsec:hsic-sel} to determine which variables are the most important and how their importance evolves through time. 

\subsection{Linear variable selection with LASSO} \label{subsec:lasso-sel}
\begin{figure}[h]
\centering
     \includegraphics[width=0.69\textwidth]{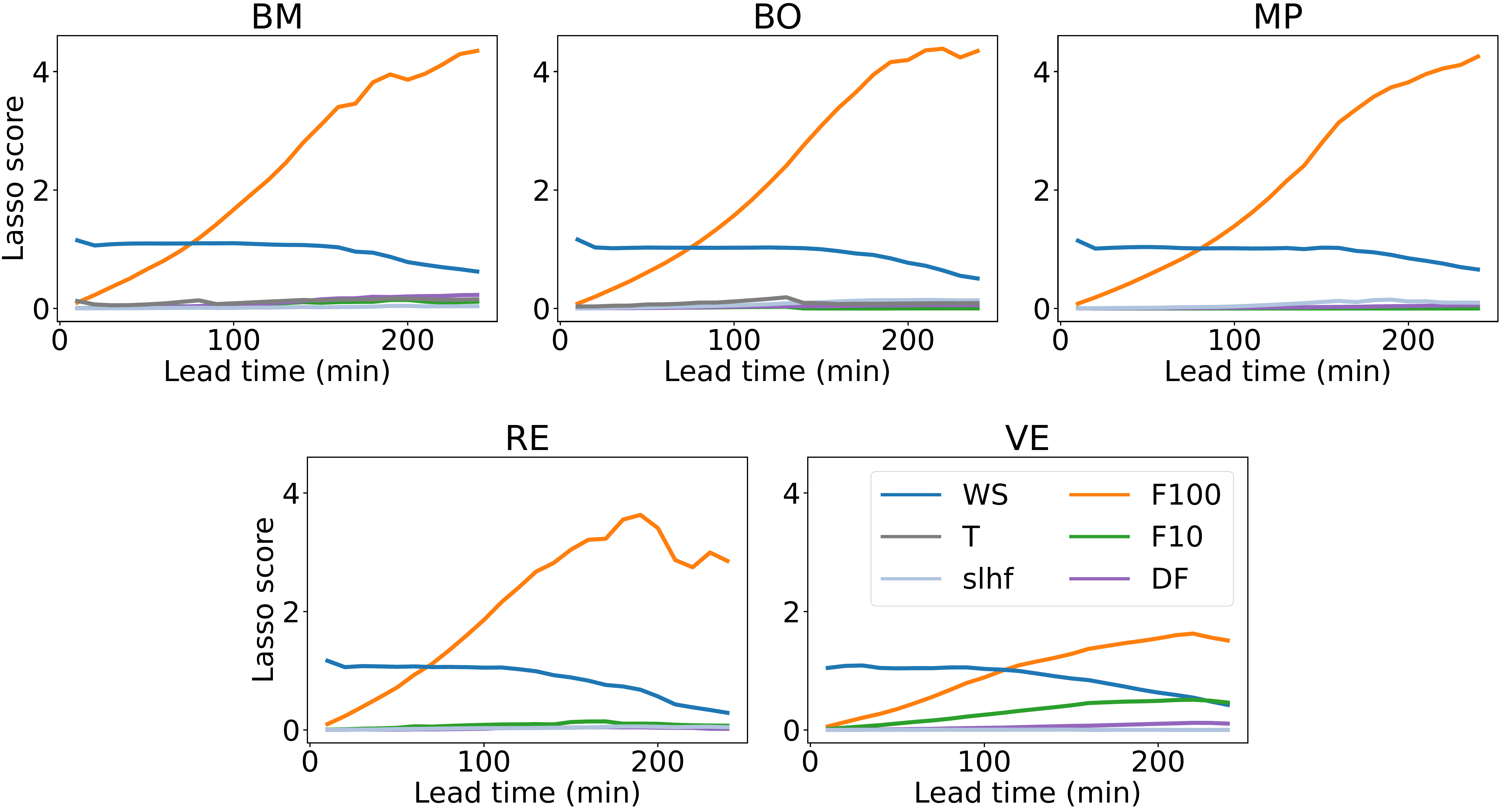}
     \caption{Linear variable selection with the LASSO (Wind speed as target)}
     \label{fig:lasso ws}
\end{figure}

\begin{figure}[h]
\centering
	\includegraphics[width=0.69\textwidth]{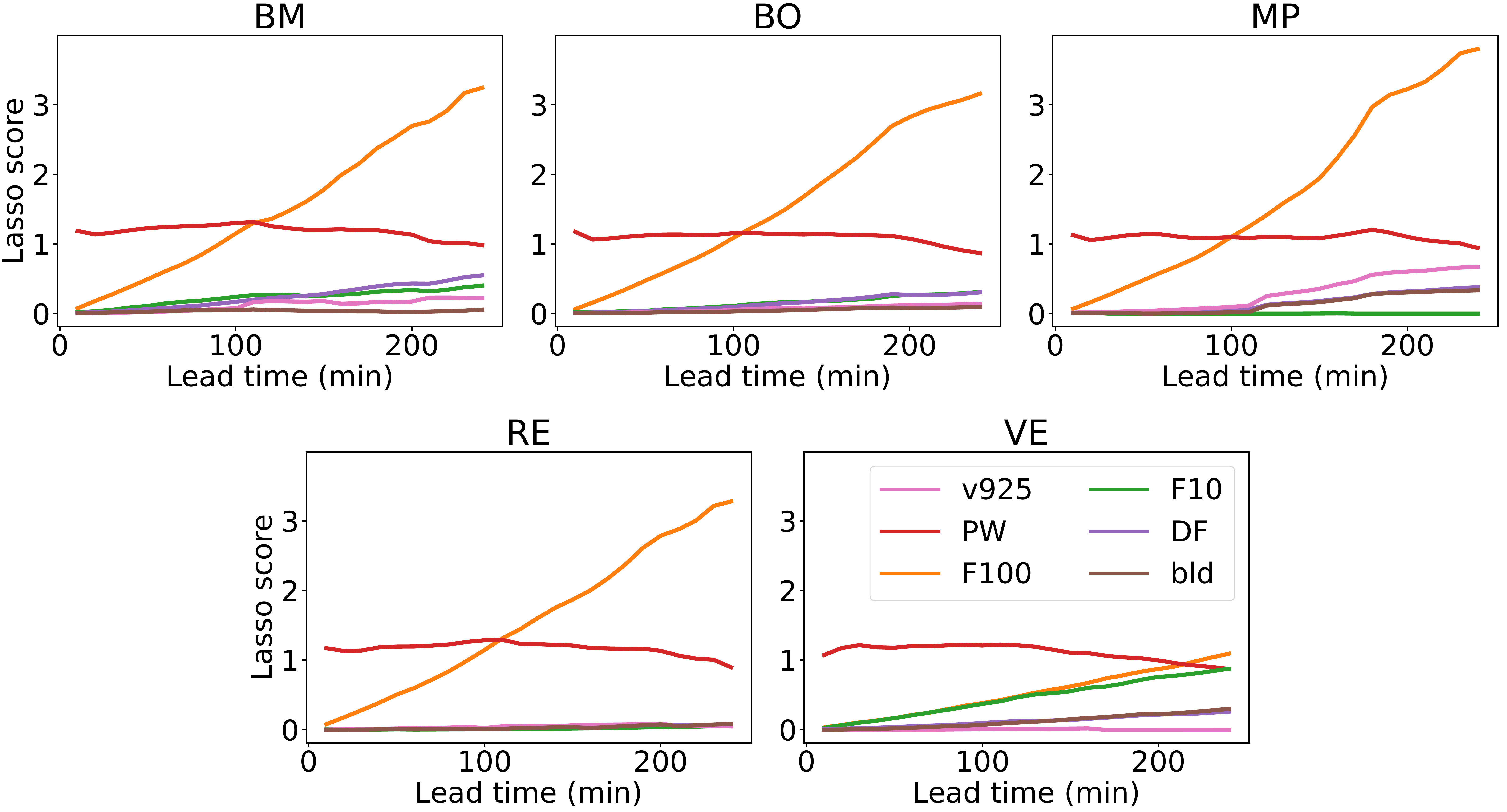}
    \caption{Linear variable selection with the LASSO (Wind power as target)}
    \label{fig:lasso pw}
\end{figure}

\noindent{\bf LASSO scores.} We now describe the computations carried out to extract a subset of relevant variables suitable for interpretation from fitted LASSO models. In practice for each data split, we validated the regularization parameter $\lambda$ on the validation set and obtained estimated LASSO coefficients. Now to reduce the number of variables we must rank them according to an importance metric based on these coefficients; we do so for each time horizon separately. So as to to avoid assigning more weight to models for which $\lambda$ was selected small, for each farm and each data split we normalize the coefficients by the absolute value of the biggest one. As opposed to the grouped variable selection performed in Section \ref{subsec:hsic-sel}, the observations through time for a given variable can be separated by the LASSO shrinking (a variable can be selected for instance at time $t_0$ but not at $t_1$). Consequently, we sum the normalized coefficients corresponding to different time instants for the same variable and in doing so, we obtain a single quantity per variable. Then we average these quantities over the data splits and call the resulting quantities LASSO scores. To sum up, at this point we have for each prediction horizon and each farm a set of such scores for each variable. Then, to select variables, we average these LASSO scores over farms.  Finally, based on these average scores, for each prediction horizon, we keep the top 6 variables.
\vspace{0.17cm}
\\
\noindent{\bf Interpretation.} For these selected variables, we plot the evolution through time of the LASSO scores in Figure \ref{fig:lasso ws} for wind speed and in Figure \ref{fig:lasso pw} for wind power. We make the following key observations: 
\begin{itemize}
\item At all locations, two variables are much more important
than the rest. The in situ observed wind speeds (WS) and the ECMWF predicted wind speed at altitude 100m
(F100) stand out for wind speed prediction. For wind power prediction, the in situ power production (PW) along with F100 are of particular importance. We can relate these results to the good performances of the LASSO for wind speed prediction in Section \ref{sec:forecasting}. Then if we look at the relative magnitudes of the coefficients, we can deduce that only using a linear combination of past local wind speeds (WS) and predicted wind speeds (F100), we can get an already good description of the future local wind speed.
\item The location VE can be singled out from the others. Indeed the predicted wind speed at altitude 10m (F10) appears, and the forecasted wind speeds (F100 and F10) take longer to take over the in situ variables, especially when predicting wind power.
This may be explained by a lesser representativity of the ECMWF forecasts for this location which may be linked to the elevation variations in the surroundings of the farm that we mentioned in Section \ref{subsec:data}.
\end{itemize}

As a concluding note, the dynamics of the local wind speed seem to be very well approximated by a simple linear model combining very few in situ variables and ECMWF ones. For predicting directly wind power however, we see in Section \ref{sec:forecasting} the results are a bit less convincing, possibly due to the nonlinear aspect of the power curve.  

\subsection{Nonlinear variable selection with HSIC} \label{subsec:hsic-sel}
\begin{figure}[h]
\centering
	\includegraphics[width=0.75\textwidth]{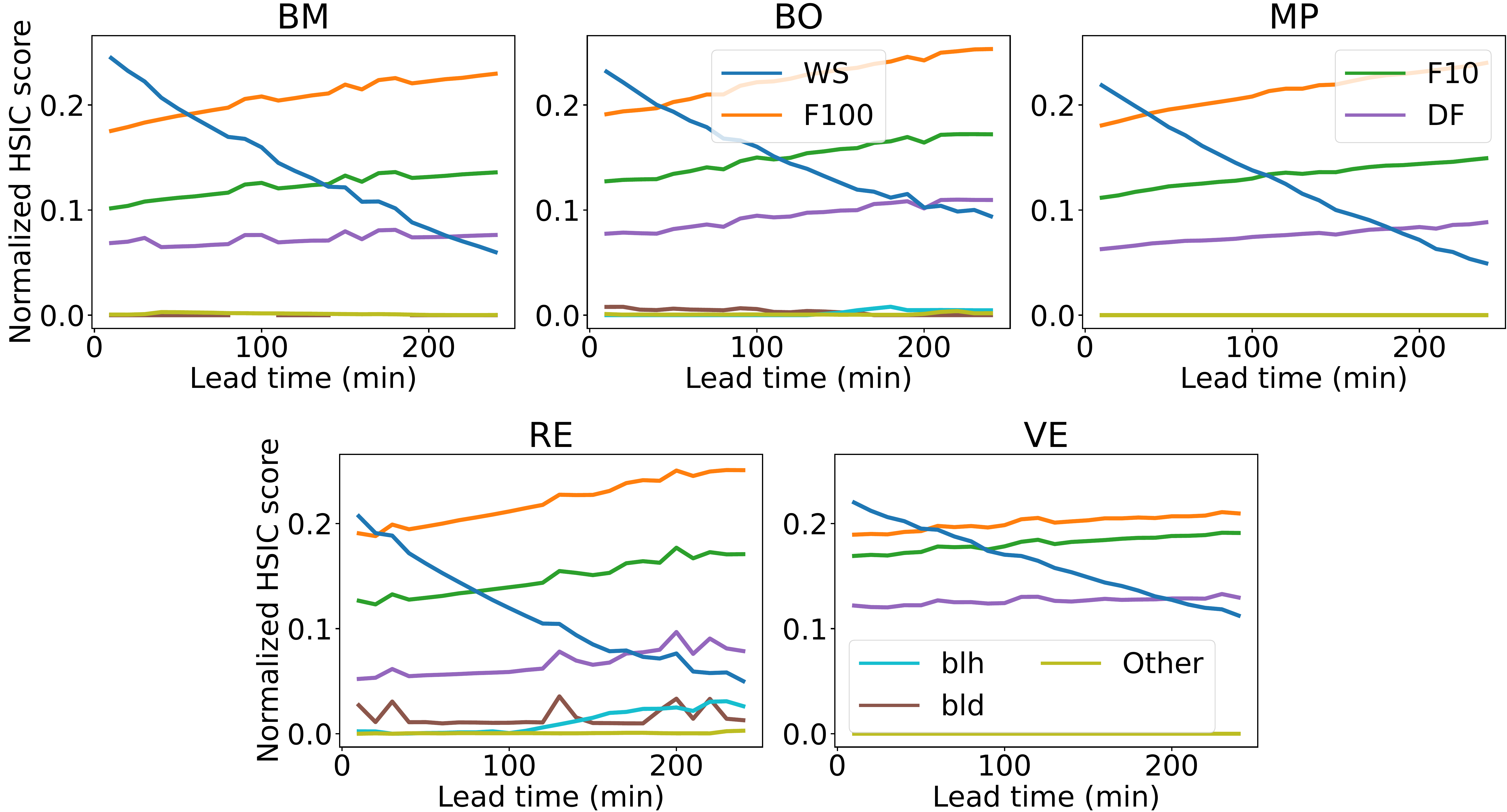}
    \caption{Nonlinear variable selection using HSIC (Wind speed as variable)}
    \label{fig:hsics-ws}
\end{figure}

\begin{figure}[h]
\centering
	\includegraphics[width=0.75\textwidth]{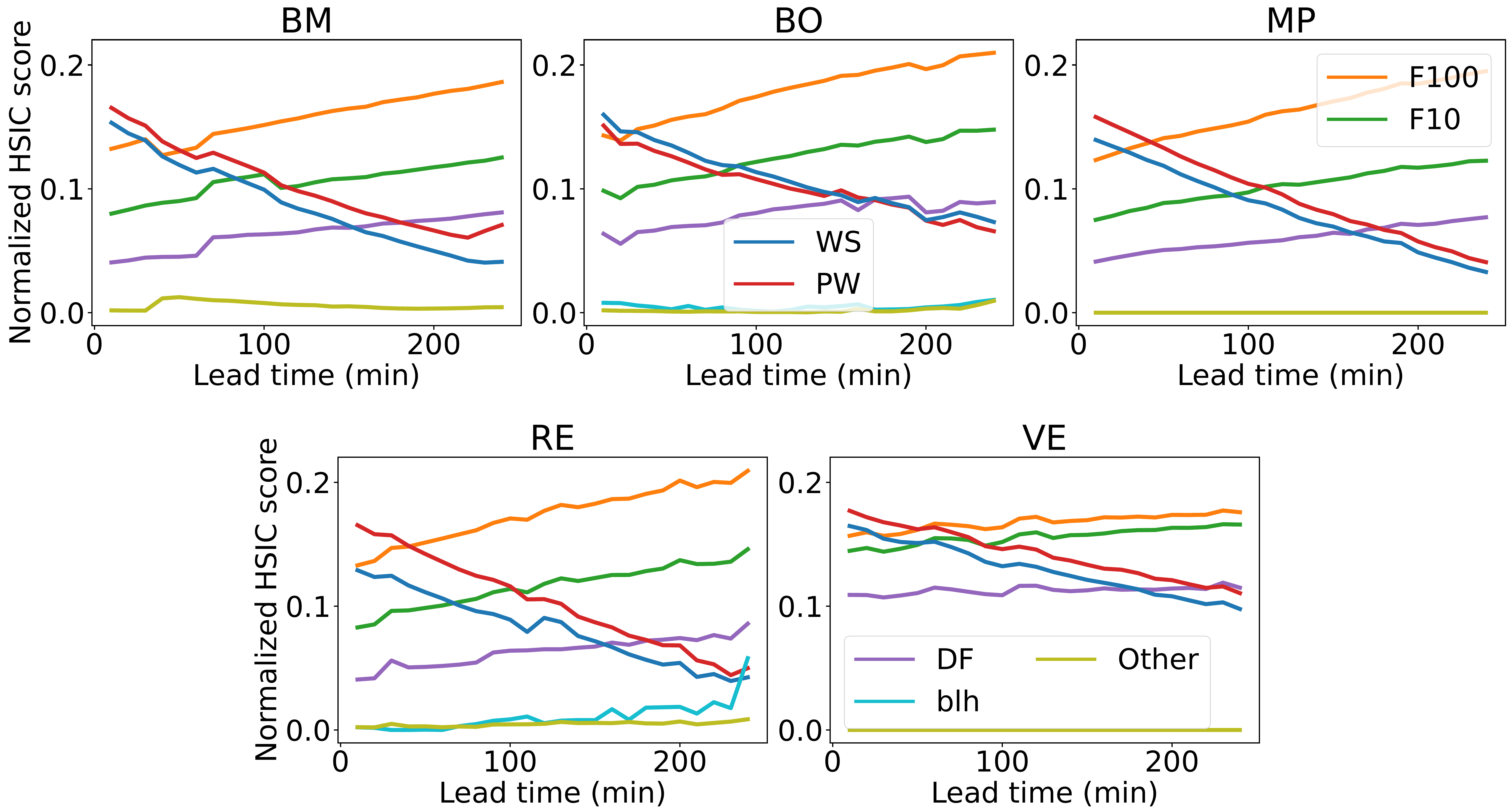}
    \caption{Nonlinear variable selection using HSIC (Wind power as target)}
    \label{fig:hsics-pw}
\end{figure}

\noindent{\bf HSIC-score.} For each farm and each split of the data we run BASHSIC on the training set until we have 5 variables left. Then for each variable we estimate the HSIC with that variable removed and normalize this value using the maximum HSIC value among these quantities. The normalized HSIC score appearing in the figures corresponds to $1$ minus this score averaged over the training sets; the higher it is, the more important the variable is. We display the results in Figure \ref{fig:hsics-ws} for wind speed and in Figure \ref{fig:hsics-pw} for wind power.
\vspace{0.17cm}

\noindent{\bf Global interpretation.} We make the following key observations: 
\begin{itemize}
\item As expected, the in situ variables are most relevant for the shortest time horizon and the ECMWF variables take progressively the lead for longer horizons. However, compared to linear feature selection, ECMWF variables take the lead faster here--between 10-50 minutes as opposed to 70-100 minutes for linear selection. The retained variables are mostly the same as the one selected by the LASSO (WS or WS and PW depending on the target) and F100. However, F10 and DF are now more systemically retained with a significant importance.
\item As for the LASSO selection, {probably due to the lesser accuracy of ECMWF forecasts for this location}, we can single out VE where the importance of the in situ variable(s) decreases much less fast than at the other farms.
\end{itemize}

\noindent{\bf Presence of DF and F10}. In comparison with linear selection, we have two more variables of interest (DF and F10). F10 describes the wind speed at lower levels and DF the wind shear near the surface. They thus bring useful information about the wind and its vertical shear, and likely help to correct deficiencies of the NWP model's description of wind at 100m. 
The fact that they appear here and not in the linear framework indicates a nonlinear relation, which is not surprising as the shear relates to the level of turbulence in the boundary layer. Additionally, the above results bring a fairly sharp answer to another question underlying this study. As the calculation of near-surface winds in NWP models involves parameterizations, they are not the most reliable output of NWP models. Consequently, one could expect that, informed about other aspects of the boundary layer and local wind realizations, a nonlinear method could capture better the relationship between the boundary layer and the near-surface winds. This is not the case: BAHSIC clearly select rather the wind variables as the best source of information. Over variable terrain (VE), wind speed at different heights (F10) are more used, suggesting that the NWP model indeed fails to accurately describe the 
wind shear. And yet variables describing the boundary layer (e.g. stratification) still remain unused or marginal. Over flat terrain, the wind speed at 100m height (F100) is the major source of information, which is positive and encouraging regarding the accuracy of NWP models.

\section{Wind speed and wind power forecasting} \label{sec:forecasting}
\begin{figure}
\centering
	\includegraphics[width=0.8\textwidth]{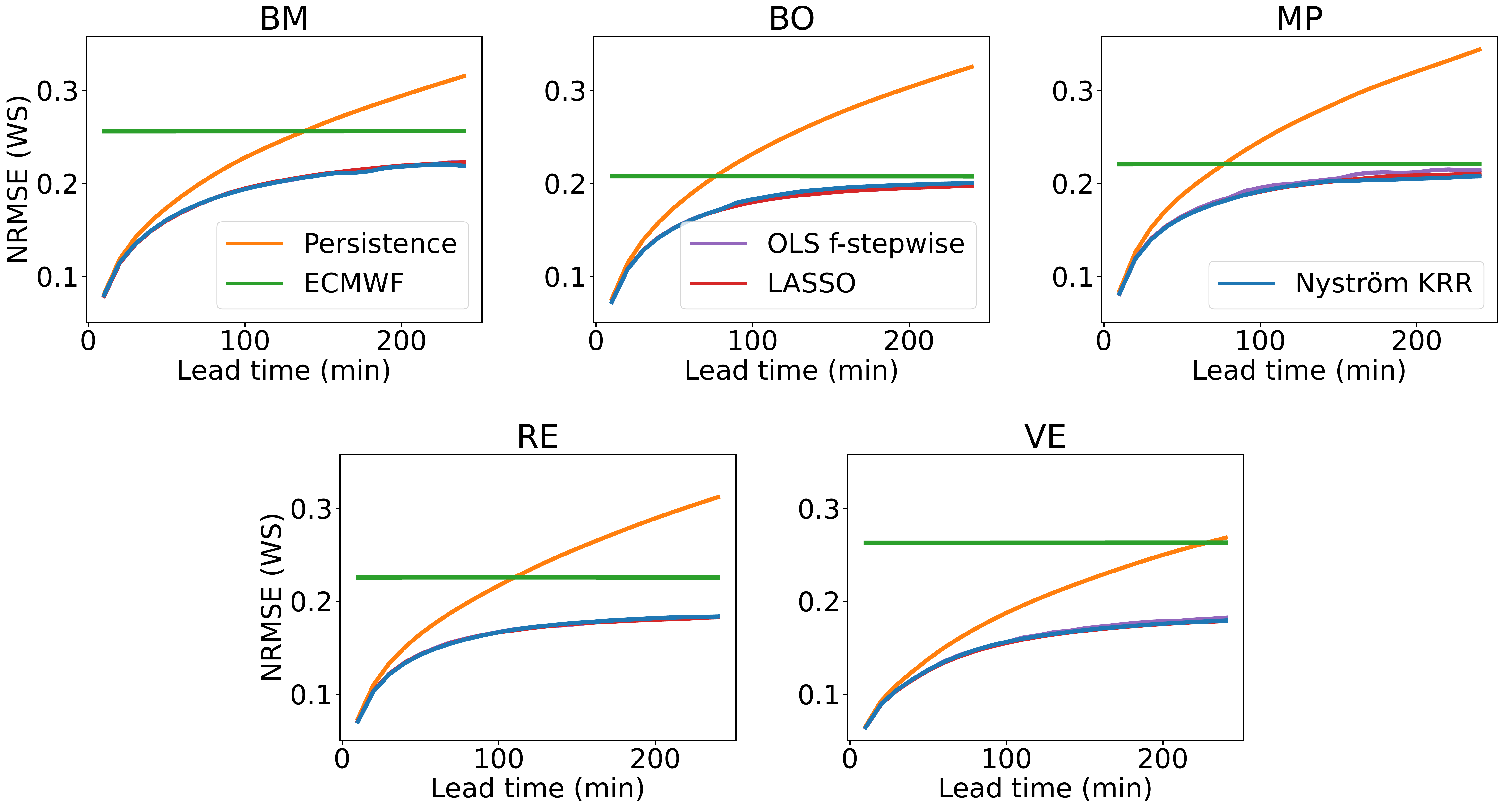}
	\parbox{\dimexpr\columnwidth-2.3cm}{
    \caption{Average NRMSE at all time horizons for the three methods performing best overall according to Table \ref{tab:ws-scores} as well as for Persistence and ECMWF (wind speed as target)}}
    \label{fig:forecasting-ws}
\end{figure}

\begin{figure}
\centering
	\includegraphics[width=0.8\textwidth]{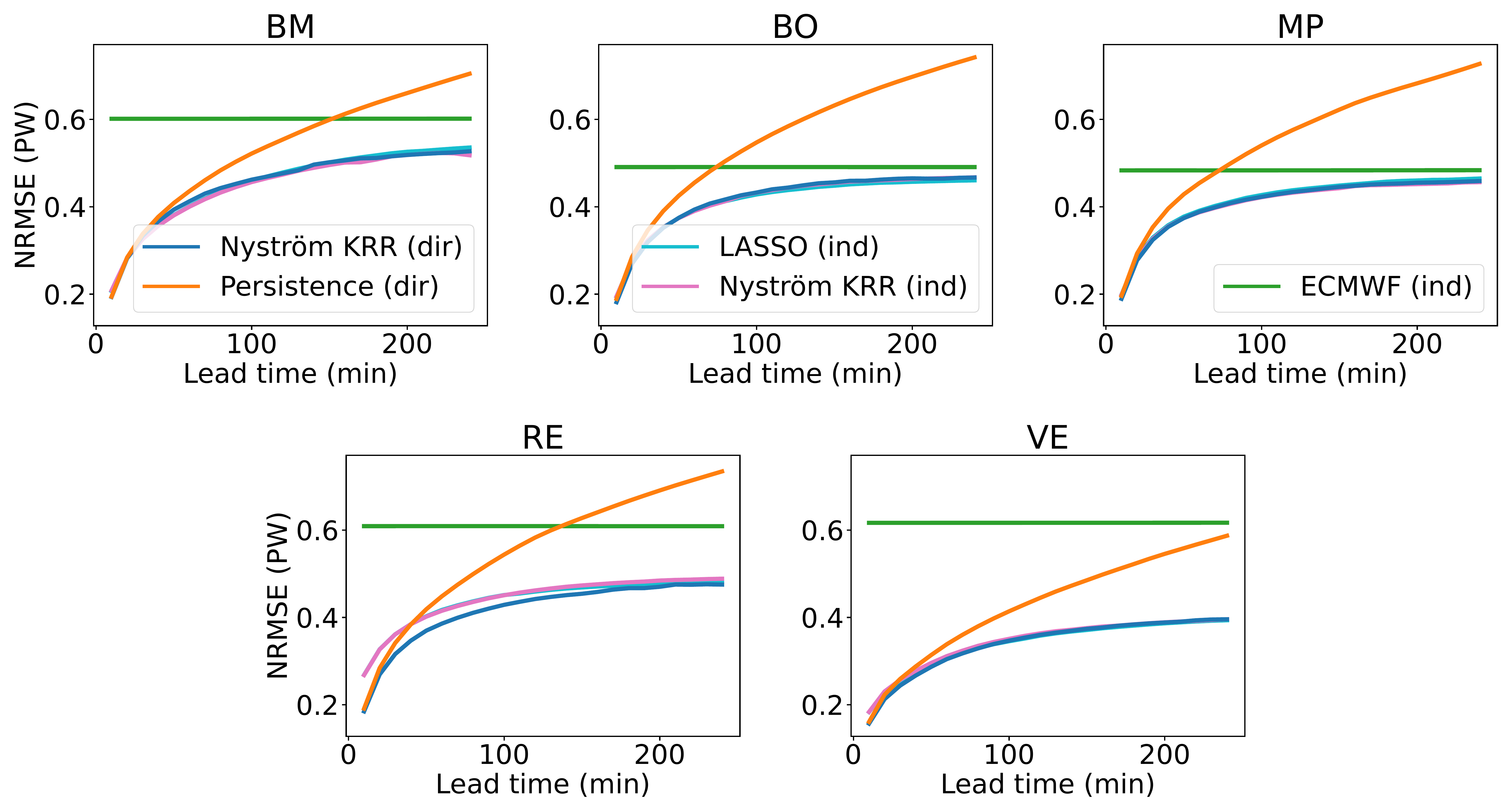}
	\parbox{\dimexpr\columnwidth-2.3cm}{
    \caption{Average NRMSE at all time horizons for the three methods performing best overall according to Table \ref{tab:pw-scores} as well as for Persistence and ECMWF (wind power as target)}}
    \label{fig:forecasting-pw}
\end{figure}

In this Section we compare several ML models for predicting
both wind speed and wind power, exploiting the variable selection techniques from the previous section. We include the two main baselines, namely persistence which predicts the last in situ observation and ECMWF which uses the F100 forecasts from the ECMWF.  

\subsection{Experimental setup} \label{subsec:expset}

\noindent{\bf Metrics.} We evaluate our results using the normalized root mean squared error (NRMSE) as in \citep{DupreAl20}. Let $(z_t)_{t=1}^n$ denote the realizations of a (scalar-valued) target variable. We define its global mean as $\bar z = \frac 1 n \sum_{t=1}^n z_t$.  Given a set of predicted values $(\widehat z_t)_{t=1}^n$, it is defined as:

$$ \text{NRMSE}: = \frac {\sqrt {\frac 1 n \sum_{t=1}^n (\widehat z_t - z_t)^2}}{\bar z}. $$

\noindent {However, in order to compare the methods over the full time span, we need to introduce a new metric. If we were to simply average NRMSEs over time, then the resulting average would not make much sense because of the difference of magnitude between the errors at the different time horizons. Therefore, we propose to compare the NRMSE at each time horizon to a specific anchor reflecting what is achievable: the NRMSE of the best predictor for this time horizon.} Let $\mathcal F$ be a given set of predictors--for instance when predicting wind speed we have $\mathcal F:=$ \{Nystr\"om KRR, LASSO, OLS f-stepwise, XG-Boost, Feedforward NN, Persistence, ECMWF\}. Given a predictor $f \in \mathcal F$, a prediction horizon $\predhor_0 \in \llbracket m \rrbracket$ and a data split $s$ (among a total of $S \in \mathbb N$ data splits), let $\text{NRMSE}_{s, \predhor_0}^{(f)}$ denote the NRMSE of predictor $f$ for the prediction horizon $\predhor_0$ on the data split $s$. We then define the average NRMSE degradation of a predictor $f_0$ (with respect to the best predictor):
\begin{equation} \label{eq:nrmse-deg}
\frac 1 {mS} \sum_{s=1}^S \sum_{\predhor_0 = 1}^\predhor \big(\text{NRMSE}_{s, \predhor_0}^{(f_0)} - \min_{f \in \mathcal F} \text{NRMSE}_{s, \predhor_0}^{(f)} \big ).
\end{equation}
The best possible value is zero as it means that over all splits and over all horizons, the predictor was the best one.
\vspace{0.17cm}
\\
\noindent{\bf Direct/indirect prediction.} When we predict wind power, we consider two prediction techniques. Either we predict directly the wind power (direct approach) or we predict the wind speed which we transform using an estimated power curve in the same fashion as in \citep{DupreAl20} (indirect approach). A theoretical power curve could be used as well, however, in this work we estimate it from the training WS and PW observations using median of nearest neighbors interpolation using $250$ neighbors.
\vspace{0.17cm}
\\
\noindent{\bf Model selection.} We follow the methodology introduced in Section \ref{subsec:methodo} and refer the reader to Section \ref{subsec:ml-details} for details on the ML methods. In practice, for each data split, the key parameters of the different methods are chosen using the validation set (the regularization parameters $\lambda$, the Gaussian
kernel's $\gamma$ for KRR, the number of variables for OLS f-stepwise, the architecture for the feedforward NN etc.). We provide the details of the considered parameters in the supplementary material.

\subsection{Comparisons over the 10 minutes - 4 hours range}

\begin{table}[h]
\begin{center}
	\begin{footnotesize}
		\begin{tabular}{c|cccccc}
		\thickhline
		\textbf{Method (average rank)} & \textbf{BM} & \textbf{BO} & \textbf{MP} & \textbf{RE} & \textbf{VE} \\ \hline 
\textbf{LASSO (1.8)} & 1.12 & \textbf{0.13} & 0.16 & \textbf{0.14} & \textbf{0.16} \\ 
Nystr\"om KRR (2.0) & \textbf{1.05} & 0.36 & \textbf{0.06} & 0.19 & 0.21 \\ 
OLS f-stepwise (2.8) & 1.1 & 0.16 & 0.47 & 0.18 & 0.34 \\ 
Feedforward NN (3.4) & 1.08 & 0.46 & 0.37 & 0.45 & 0.66 \\ 
XG Boost (5.0) & 1.81 & 1.25 & 1.63 & 0.74 & 0.97 \\ 
ECMWF (6.4) & 7.81 & 3.63 & 3.75 & 6.66 & 11.37 \\ 
Persistence (6.6) & 5.59 & 6.71 & 7.02 & 6.7 & 4.5 \\ 
		\end{tabular}
		\end{footnotesize}
		\parbox{\dimexpr\columnwidth-2.3cm}{
       \caption{Average NRMSE degradation w.r.t. best predictor for wind speed prediction ($\times 10^{-2}$)}}
       \label{tab:ws-scores}
       \end{center}
\end{table}

\begin{table}[h]
\begin{center}
	\begin{footnotesize}
		\begin{tabular}{c|c|ccccc}
		\thickhline
		 \textbf{Type} & \textbf{Method (average rank)} & \textbf{BM} & \textbf{BO} & \textbf{MP} & \textbf{RE} & \textbf{VE} \\ \hline 
\textbf{Direct} & \textbf{Nystr\"om KRR (2.4)} & 4.01 & 1.2 & 0.63 & \textbf{0.46} & \textbf{0.7} \\ 
Indirect & Nystr\"om KRR (3.0) & \textbf{3.54} & 1.19 & \textbf{0.55} & 2.97 & 1.2 \\ 
Indirect & LASSO (3.4) & 4.04 & \textbf{0.8} & 1.08 & 2.72 & 0.87 \\ 
Direct & Feedforward NN (4.0) & 3.56 & 1.71 & 2.29 & 1.08 & 1.67 \\ 
Indirect & OLS f-stepwise (4.4) & 3.7 & 0.87 & 1.53 & 3.42 & 1.61 \\ 
Direct & XG Boost (direct) (5.2) & 4.62 & 2.39 & 1.5 & 1.8 & 1.8 \\ 
Direct & OLS f-stepwise (6.6) & 4.15 & 3.01 & 3.56 & 2.42 & 3.0 \\ 
Direct & LASSO (direct) (7.0) & 4.91 & 3.15 & 3.46 & 2.54 & 2.46 \\ 
Direct & Persistence (9.4) & 12.08 & 15.48 & 14.91 & 14.25 & 9.88 \\ 
Indirect & ECMWF (10.2) & 18.93 & 8.66 & 8.02 & 19.74 & 28.53 \\ 
Indirect & Persistence (10.4) & 13.21 & 15.89 & 15.22 & 16.39 & 10.86 \\ 
		\end{tabular}
		\end{footnotesize}
		\parbox{\dimexpr\columnwidth-2.3cm}{
       \caption{Average NRMSE degradation w.r.t. best predictor for wind power
       prediction ($\times 10^{-2}$)}}
       \label{tab:pw-scores}
       \end{center}
\end{table}

\noindent{\bf Overall efficiency of ML models.} From a general perspective, our experiments show that combining a NWP model's outputs with local observations is very beneficial for predicting both wind speed and wind power at all the time horizons considered. {To that end, Figure \ref{fig:forecasting-ws} displays the evolution of the NRSME for the two baselines (persistence and ECMWF) as well as for the three ML methods which performed best for wind speed prediction in Table \ref{tab:ws-scores} while Figure \ref{fig:forecasting-pw} displays the same for wind power prediction; the displayed ML methods being the top three ones from Table \ref{tab:pw-scores}.} For three farms (VE and to a lesser extend, BM and RE), even after four hours the improvement over ECMWF is still quite large. For BO and MP it becomes less important, yet still present. The improvement can be quite dramatic for very short horizons (first 100 minutes or so), and a bit less important for longer time horizons. This is probably linked to the representativity of the NWP model's outputs which depends on the location.
\vspace{0.17cm}
\\
\noindent{\bf Quantitative comparison.} We now use the NRMSE degradation w.r.t. the best predictor (Equation \ref{eq:nrmse-deg}) to compare the methods. The results are displayed in Table \ref{tab:ws-scores} (WS as target) and in Table \ref{tab:pw-scores} (PW as target). On the one hand, for wind speed prediction, the LASSO is the best ranked method. Relating this to the results from Section \ref{subsec:lasso-sel}, it shows that the dynamics of the wind speed can be very well described by a linear combination of few ECMWF and local variables (essentially past local wind speeds and forecasted wind speeds). It suggests that the important nonlinear dynamics are overall well captured in the ECMWF variables. 
On the other hand, it seems better to predict directly wind power and do so using the Nyström KRR. This is not surprising, as the power curve is a nonlinear function and so we expected the linear methods to struggle in direct prediction. Moreover, in direct prediction, we implicitly include the power curve into the learning problem. This is advantageous since for instance a model trained to predict wind speed first will be very eager to forecast well high values (failing to do so would incur a high error term). However to predict wind power, producing accurate forecasts for higher wind speeds is less critical, since in the power curve, the actual wind power as a function of the wind speed is thresholded. 

{
We note that the feedforward NN does not beat indirect prediction with the LASSO. This suggests that the higher expressiveness of NNs beyond the ability to infer the nonlinearity of the power curve is not needed. The difference of performance with direct Nyström KRR is imputable to the optimization error and variability implied by non-convexity of NNs whereas for Nyström KRR the optimization error is close to non-existent thanks to the closed-form solution. XG-Boost also does not perform well, this can be explained by the use of time series as features: these are very correlated and high dimensional which can make tree-based models unstable \citep{GregoruttiAl17importance}. Doing some more work on feature pre-processing should improve the results. 
}
{
\subsection{Zoom on 10 minutes and 1 hour ahead forecasting} \label{subsec:1060}

\begin{table}[h]
\begin{center}
	\begin{footnotesize}
        \centering
		\begin{tabular}{c|c|cccccc}
		\thickhline
		\textbf{Horizon} & \textbf{Method (average rank)} & \textbf{BM} & \textbf{BO} & \textbf{MP} & \textbf{RE} & \textbf{VE} \\ \hline 
10 min & \textbf{Nystr\"om KRR (1.8)} & 8.0 & \textbf{7.25} & 8.15 & \textbf{7.08} & 6.44 \\ 
& LASSO (2.0) & \textbf{7.91} & 7.25 & 8.15 & 7.12 & \textbf{6.43} \\ 
& OLS f-stepwise (2.2) & 7.92 & 7.25 & \textbf{8.14} & 7.09 & 6.44 \\ 
& Persistence (4.4) & 8.11 & 7.55 & 8.43 & 7.39 & 6.56 \\ 
& Feedforward NN (4.8) & 8.26 & 7.55 & 8.33 & 7.37 & 6.8 \\ 
& XG Boost (5.8) & 8.72 & 8.26 & 9.57 & 7.61 & 6.77 \\ 
& ECMWF (7.0) & 25.61 & 20.78 & 22.06 & 22.57 & 26.3 \\ 
\hline 
1 hour & \textbf{Nystr\"om KRR (1.6)} & 17.01 & \textbf{16.03} & \textbf{17.11} & \textbf{14.94} & 13.48 \\ 
& LASSO (1.8) & 16.94 & 16.04 & 17.13 & 14.95 & \textbf{13.4} \\ 
& OLS f-stepwise (2.6) & \textbf{16.92} & 16.06 & 17.3 & 15.02 & 13.52 \\ 
& Feedforward NN (4.0) & 17.27 & 16.27 & 17.32 & 15.17 & 13.72 \\ 
& XG Boost (5.0) & 17.69 & 17.19 & 18.77 & 15.66 & 14.16 \\ 
& Persistence (6.0) & 18.68 & 18.81 & 20.09 & 17.73 & 15.0 \\ 
& ECMWF (7.0) & 25.61 & 20.78 & 22.06 & 22.58 & 26.3 \\ 
		\end{tabular}
		\end{footnotesize}
		\parbox{\dimexpr\columnwidth-2.3cm}{
       \caption{{Average NRMSE for 10 minutes and 1 hour ahead wind speed prediction ($\times 10^{-2}$)}}}
       \label{tab:ws-scores1060}
    \end{center}
\end{table}
\begin{table}[h]
\begin{center}
	\begin{footnotesize}
		\begin{tabular}{c|c|c|ccccc}
		\thickhline
		 \textbf{Horizon} & \textbf{Type} & \textbf{Method (average rank)} & \textbf{BM} & \textbf{BO} & \textbf{MP} & \textbf{RE} & \textbf{VE} \\ \hline 
10 min & \textbf{Direct} & \textbf{Nystr\"om KRR(1.4)} & 19.36 & \textbf{18.16} & \textbf{18.94} & \textbf{18.49} & \textbf{15.7} \\ 
& Direct & OLS f-stepwise (2.0) & \textbf{19.04} & 18.2 & 18.96 & 18.53 & 15.8 \\ 
& Direct & LASSO (2.6) & 19.06 & 18.22 & 19.01 & 18.56 & 15.78 \\ 
& Direct & Persistence (4.4) & 19.44 & 18.92 & 19.63 & 19.13 & 16.09 \\ 
& Direct & Feedforward NN (4.6) & 19.73 & 18.81 & 19.56 & 19.3 & 16.3 \\ 
& Direct & XG Boost (6.0) & 19.9 & 19.41 & 19.77 & 19.38 & 16.59 \\ 
& Indirect & LASSO (7.6) & 20.73 & 19.48 & 19.91 & 26.84 & 18.34 \\ 
& Indirect & Nystr\"om KRR (8.0) & 20.86 & 19.51 & 19.87 & 26.8 & 18.36 \\ 
& Indirect & OLS f-stepwise (8.4) & 20.73 & 19.48 & 19.89 & 26.87 & 18.39 \\ 
& Indirect & Persistence (10.0) & 21.67 & 20.41 & 20.77 & 27.32 & 18.8 \\ 
& Indirect & ECMWF (11.0) & 60.15 & 49.09 & 48.34 & 60.91 & 61.66 \\ 
\hline
1 hour & \textbf{Indirect} & \textbf{Nystr\"om KRR (2.8)} & 40.02 & \textbf{39.04} & \textbf{38.74} & 41.53 & 31.15 \\ 
& Direct & Nystr\"om KRR (3.0) & 41.26 & 39.36 & 38.85 & \textbf{38.59} & \textbf{30.46} \\ 
& Indirect & LASSO (3.4) & 40.11 & 39.07 & 39.1 & 41.72 & 30.8 \\ 
& Indirect & OLS f-stepwise (4.6) & \textbf{39.97} & 39.16 & 39.37 & 42.2 & 31.36 \\ 
& Direct & Feedforward NN (4.8) & 40.51 & 39.64 & 39.58 & 39.44 & 31.16 \\ 
& Direct & LASSO (5.2) & 40.39 & 39.93 & 40.06 & 39.83 & 31.07 \\ 
& Direct & OLS f-stepwise (5.6) & 40.37 & 39.82 & 40.26 & 39.67 & 31.53 \\ 
& Direct & XG Boost (6.6) & 41.89 & 40.43 & 39.22 & 39.99 & 31.81 \\ 
& Direct & Persistence (9.0) & 43.58 & 45.52 & 45.44 & 44.84 & 33.88 \\ 
& Indirect & Persistence (10.0) & 44.68 & 46.01 & 45.99 & 47.36 & 35.08 \\ 
& Indirect & ECMWF (11.0) & 60.15 & 49.1 & 48.34 & 60.91 & 61.67 \\ 
		\end{tabular}
		\end{footnotesize}
		\parbox{\dimexpr\columnwidth-2.3cm}{
       \caption{{Average NRMSE for 10 minutes and 1 hour ahead wind power prediction ($\times 10^{-2}$)}}}
       \label{tab:pw-scores1060}
       \end{center}
\end{table}

We now propose to zoom in on on two time horizons of particular interest: 10 minutes and 1 hour ahead. We display the raw NRSMEs in Table \ref{tab:ws-scores1060} for WS and in Table \ref{tab:pw-scores1060} for PW. 

For 10 minutes ahead prediction, persistence is unsurprisingly very efficient even though small yet significant improvements are already obtained by exploiting also ECMWF information with ML. For both the prediction of WS and PW, all three methods which beat persistence reach similar scores. For WS, these are the same as those performing best overall in Table \ref{tab:ws-scores}: Nyström KRR, LASSO and OLS f-stepwise. However, for PW, these are Nyström KRR (direct), OLS f-stepwise (direct), LASSO (direct). The first is the leading method in Table \ref{tab:ws-scores}, but the other two are not. We explained their poor performance by the nonlinearity of the power curve which they cannot capture. Nevertheless for the very short-term, this is not an issue. This confirms our findings from Section \ref{sec:varsel}: for 10 minutes ahead prediction, the last observed wind power is the most crucial information. 

For 1 hour ahead prediction, the rankings of ML methods for both wind speed (Table \ref{tab:ws-scores1060}) and wind power (Table \ref{tab:pw-scores1060}) almost perfectly match the rankings of methods on the whole time span (Table \ref{tab:ws-scores} for WS and Table \ref{tab:pw-scores} for PW). 

Overall, this analysis confirms that for both the very short term and the longer term Nyström KRR is a safe choice for wind speed and wind power prediction. For the latter the direct approach with this method should be preferred.

\subsection{Computational times} \label{subsec:comptimes}

\begin{table}[h]
    \begin{center}
	\begin{footnotesize}
		\begin{tabular}{c|c|cc}
		\thickhline
		\textbf{Task} & \textbf{Method} & \textbf{Fit time (s)} & \textbf{Predict time (s)} \\
		\hline
		Selection & Nystr\"om BAHSIC & 74.61 (73.38, 89.00) & - \\
		Selection \& regression & LASSO & 1.39 (0.01, 5.03) & 0.039 (0.036, 0.044) \\
		Selection \& regression & OLS f-stepwise & 3.58 (2.92, 4.30) & 0.037 (0.037, 0.043) \\
		Regression & Nystr\"om KRR & 0.451 (0.414, 0.563) & 0.332 (0.300, 0.433) \\
		Regression & XgBoost & 0.450 (0.405, 0.548) & 0.058 (0.053, 0.070) \\
		Regression & Feedforward NN & 75.84 (72.70, 77.27) & 0.029 (0.028, 0.031) \\ 
		\end{tabular}
		\end{footnotesize}
		\\
		\parbox{\dimexpr\columnwidth-2.3cm}{
		\caption{{Median (10 \% quantile, 90 \% quantile) of fit and predict computational times on laptop for direct wind power prediction on BO farm.}}}
       \label{tab:comptimes}
	\end{center}
\end{table}
}

We now address the practical concern of computational times. To that end, we measure the time on a laptop to fit the different procedures and to produce the corresponding forecasts. We do so only for one wind farm (BO). We draw randomly $50$ pairs containing a split from the dataset (see Section \ref{subsec:data-methodo}) and a parameter configuration among the ones we used. Then we time the procedures using these pairs. We display the median as well as the 10\% and 90\% quantiles of the obtained computational times in Table \ref{tab:comptimes}. To put these computational times into perspective, on the one hand the regression models have extra parameters to tune, and therefore many configurations must be tested. On the other hand Nyström BAHSIC seems expensive but no such tuning must be performed (it eliminates the variables gradually, therefore the ranking can be used to include more or less features afterwards). 

\section{Conclusion} \label{sec:conclusion}

We showed through experiments on several wind farms that we can improve very significantly short-term local forecasts of both wind speed and wind power by combining statistically a NWP model's outputs with local observations. To better understand how, we studied in details the evolution of the variables' importance using two metrics, a linear one based on LASSO coefficients and a nonlinear one using HSIC. Our global conclusion is that NWP wind variables are a very relevant source of information to complement local observations, even for the very short-term. To forecast wind speed, a parsimonious linear combination of NWP and local variables (with the LASSO) yielded the best result. While to forecast wind power, direct prediction (no power
curve involved) with a nonlinear method (Nystr\"om KRR) using a few variables (selected with BAHSIC) is preferable. Beyond the ability to capture the nonlinearity of the power curve, it seems unnecessary to use more complex models which hints that NWP model's outputs describe sufficiently the other nonlinear dynamics involved. For future work, assessing the variability of the predictions, for instance by predicting conditional quantiles \citep{KoenkerHallock01} which would inform us on the expected
distribution of the predictions. This could help mitigate the intermittent effects of wind power production further. 
%
%
%

\section*{Acknowledgments}

This work received support from the Télécom Paris research chair on Data Science
and Artificial Intelligence for Digitalized Industry and Services (DSAIDIS). It was conducted as part of the Energy4Climate Interdisciplinary Center (E4C) of Institut Polytechnique de Paris
and Ecole des Ponts ParisTech. It has received support as well from the 3rd
Programme d’Investissements d’Avenir [ANR-18-EUR-0006-02] and from the
Foundation of Ecole polytechnique (Chaire “Défis Technologiques pour une Énergie
Responsable”). The authors would like to thank the company Z\'ephyr ENR, and
especially its CEO Christian Briard, for providing the data from its wind farms.

\bibliographystyle{plainnat}
\bibliography{biblio}

\begin{thebibliography}{35}
\providecommand{\natexlab}[1]{#1}
\providecommand{\url}[1]{\texttt{#1}}
\expandafter\ifx\csname urlstyle\endcsname\relax
  \providecommand{\doi}[1]{doi: #1}\else
  \providecommand{\doi}{doi: \begingroup \urlstyle{rm}\Url}\fi

\bibitem[Baars and Mass(2005)]{BaarsMass2005}
J.A. Baars and C.F. Mass.
\newblock {Performance of National Weather Service forecasts compared to
  operational, consensus and weighted model output statistics}.
\newblock \emph{Wea. Forecast.}, 20:\penalty0 1034--1047, 2005.

\bibitem[Bauer et~al.(2015)Bauer, Thorpe, and Brunet]{BTB15}
P.~Bauer, A.~Thorpe, and G.~Brunet.
\newblock {The quiet revolution of numerical weather prediction}.
\newblock \emph{Nature}, 525:\penalty0 47--55, 2015.

\bibitem[Beck and Teboulle(2009)]{BeckTeboulle09}
Amir Beck and Marc Teboulle.
\newblock A fast iterative shrinkage-thresholding algorithm for linear inverse
  problems.
\newblock \emph{SIAM Journal on Imaging Sciences}, 2\penalty0 (1):\penalty0
  183--202, 2009.

\bibitem[Chen and Guestrin(2016)]{ChenGuestrin16}
Tianqi Chen and Carlos Guestrin.
\newblock Xgboost: A scalable tree boosting system.
\newblock In \emph{Proceedings of the 22nd ACM SIGKDD International Conference
  on Knowledge Discovery and Data Mining}, pages 785--794, 2016.

\bibitem[Drineas and Mahoney~W.(2005)]{DrineasMahoney05}
Petros Drineas and Michael Mahoney~W.
\newblock On the nystrom method for approximating a gram matrix for improved
  kernel-based learning.
\newblock \emph{Journal of Machine Learning Research}, 6:\penalty0 2153--2175,
  2005.

\bibitem[Dupr\'e et~al.(2020)Dupr\'e, Drobinski, Alonzo, Badosa, Briard, and
  Plougonven]{DupreAl20}
Aurore Dupr\'e, Philippe Drobinski, Bastien Alonzo, Jordi Badosa, Christian
  Briard, and Riwal Plougonven.
\newblock Sub-hourly forecasting of wind speed and wind energy.
\newblock \emph{Renewable Energy}, 145:\penalty0 2373--2379, 2020.

\bibitem[Dupr{\'e} et~al.(2020)Dupr{\'e}, Drobinski, Badosa, Briard, and
  Tankov]{dupre2020economic}
Aurore Dupr{\'e}, Philippe Drobinski, Jordi Badosa, Christian Briard, and Peter
  Tankov.
\newblock The economic value of wind energy nowcasting.
\newblock \emph{Energies}, 13\penalty0 (20):\penalty0 5266, 2020.

\bibitem[Friedman(2001)]{Friedman01}
Jerome~H. Friedman.
\newblock Greedy function approximation: A gradient boosting machine.
\newblock \emph{The Annals of Statistics}, 29\penalty0 (5):\penalty0 1189 --
  1232, 2001.

\bibitem[Glahn and Lowry(1972)]{GL72}
H.R. Glahn and D.A. Lowry.
\newblock {The use of model output statistics (MOS) in objective weather
  forecasting}.
\newblock \emph{J. App. Meteor.}, 11:\penalty0 1203--1211, 1972.

\bibitem[Goutham et~al.(2021)Goutham, Alonzo, Dupr\'e, Plougonven, Doctors,
  Liao, Mougeot, Fischer, and Drobinski]{Gouthametal2021}
N.~Goutham, B.~Alonzo, A.~Dupr\'e, R.~Plougonven, R.~Doctors, L.~Liao,
  M.~Mougeot, A;~Fischer, and P.~Drobinski.
\newblock Using machine-learning methods to improve surface wind speed from the
  outputs of a numerical weather prediction model.
\newblock \emph{Boundary Layer Meteorology}, 2021.

\bibitem[Gregorutti et~al.(2017)Gregorutti, Michel, and
  Saint-Pierre]{GregoruttiAl17importance}
Baptiste Gregorutti, Bertrand Michel, and Philippe Saint-Pierre.
\newblock Correlation and variable importance in random forests.
\newblock \emph{Statistics and Computing}, 27:\penalty0 659--678, 2017.

\bibitem[Gretton et~al.(2005)Gretton, Bousquet, Smola, and
  Sch{\"o}lkopf]{GrettonHsic05}
Arthur Gretton, Olivier Bousquet, Alex Smola, and Bernhard Sch{\"o}lkopf.
\newblock Measuring statistical dependence with hilbert-schmidt norms.
\newblock In \emph{Algorithmic Learning Theory}, pages 63--77. Springer Berlin
  Heidelberg, 2005.

\bibitem[Gretton et~al.(2008)Gretton, Fukumizu, Teo, Song, Sch\"{o}lkopf, and
  Smola]{GrettonAl08}
Arthur Gretton, Kenji Fukumizu, Choon Teo, Le~Song, Bernhard Sch\"{o}lkopf, and
  Alex Smola.
\newblock A kernel statistical test of independence.
\newblock In \emph{Advances in Neural Information Processing Systems (NIPS)},
  volume~20, 2008.

\bibitem[Han et~al.(2022)Han, Mi, Shen, Cai, Liu, Li, and Xu]{HanAl22newdeep}
Yan Han, Lihua Mi, Lian Shen, C.S. Cai, Yuchen Liu, Kai Li, and Guoji Xu.
\newblock A short-term wind speed prediction method utilizing novel hybrid deep
  learning algorithms to correct numerical weather forecasting.
\newblock \emph{Applied Energy}, 312, 2022.

\bibitem[Hastie et~al.(2001)Hastie, Friedman, and Tibshirani]{HastieAl01}
Trevor Hastie, Jerome~H. Friedman, and Robert Tibshirani.
\newblock \emph{The Elements of Statistical Learning}.
\newblock Springer, 2001.

\bibitem[Hoolohan et~al.(2018)Hoolohan, Tomlin, and Cockerill]{HoolohanAl18gps}
Victoria Hoolohan, Alison~S. Tomlin, and Timothy Cockerill.
\newblock Improved near surface wind speed predictions using gaussian process
  regression combined with numerical weather predictions and observed
  meteorological data.
\newblock \emph{Renewable Energy}, 126:\penalty0 1043--1054, 2018.

\bibitem[{IEA}(2018)]{IEA_Renewables2018}
{IEA}.
\newblock Renewables information.
\newblock \emph{International Energy Agency}, page~12, 2018.

\bibitem[{IEA}(2021)]{IEARenewables2021}
{IEA}.
\newblock Renewables: Analysis and forecast to 2026.
\newblock \emph{International Energy Agency}, page 175, 2021.

\bibitem[Kalnay(2003)]{Kalnay2003}
Eugenia Kalnay.
\newblock \emph{Atmospheric modeling, data assimilation and predictability}.
\newblock Cambridge University Press, 2003.

\bibitem[Koenker and Hallock(2001)]{KoenkerHallock01}
Roger Koenker and Kevin~F. Hallock.
\newblock Quantile regression.
\newblock \emph{Journal of Economic Perspectives}, 15\penalty0 (4):\penalty0
  143--156, December 2001.

\bibitem[Liu et~al.(2019)Liu, Chen, Lv, Wu, and Liu]{Liuetal2019}
Hui Liu, Chao Chen, Xinwei Lv, Xing Wu, and Min Liu.
\newblock Deterministic wind energy forecasting: {A} review of intelligent
  predictors and auxiliary methods.
\newblock \emph{Energy Conversion and Management}, 195:\penalty0 328--345,
  September 2019.

\bibitem[Masson-Delmotte et~al.(2021)Masson-Delmotte, Zhai, Pirani, Connors,
  P\'ean, Berger, Caud, Chen, Goldfarb, Gomis, Huang, Leitzell, Lonnoy,
  Matthews, Maycock, Waterfield, Yelek\c{c}i, Yu, and (eds.)]{AR6SP}
V.~Masson-Delmotte, P.~Zhai, A.~Pirani, S.L. Connors, C.~P\'ean, S.~Berger,
  N.~Caud, Y.~Chen, L.~Goldfarb, M.~I. Gomis, M.~Huang, K.~Leitzell, E.~Lonnoy,
  J.B.R. Matthews, T.~K. Maycock, T.~Waterfield, O.~Yelek\c{c}i, R.~Yu, and
  B.~Zhou (eds.).
\newblock {IPCC, 2021: Sumarry for Policymakers}.
\newblock In \emph{Climate Change 2021: The Physical Science Basis.
  Contribution of Working Group I to the Sixth Assessment Report of the
  Intergovernmental Panel on Climate Change}, pages 1--42, 2021.

\bibitem[Okumus and Dinler(2016)]{OkumusDinler2016}
Inci Okumus and Ali Dinler.
\newblock Current status of wind energy forecasting and a hybrid method for
  hourly predictions.
\newblock \emph{Energy Conversion and Management}, 123:\penalty0 362--371,
  September 2016.

\bibitem[Pedregosa et~al.(2011)Pedregosa, Varoquaux, Gramfort, Michel, Thirion,
  Grisel, Blondel, Prettenhofer, Weiss, Dubourg, et~al.]{Scikit11}
Fabian Pedregosa, Ga{\"e}l Varoquaux, Alexandre Gramfort, Vincent Michel,
  Bertrand Thirion, Olivier Grisel, Mathieu Blondel, Peter Prettenhofer, Ron
  Weiss, Vincent Dubourg, et~al.
\newblock Scikit-learn: Machine learning in python.
\newblock \emph{Journal of machine learning research}, 12\penalty0
  (Oct):\penalty0 2825--2830, 2011.

\bibitem[Rudi et~al.(2015)Rudi, Camoriano, and Rosasco]{RudiAl15}
Alessandro Rudi, Raffaello Camoriano, and Lorenzo Rosasco.
\newblock Less is more: Nystr\"{o}m computational regularization.
\newblock In \emph{Advances in Neural Information Processing Systems (NIPS)},
  pages 1657--1665, 2015.

\bibitem[Sch\"olkopf and Smola(2002)]{ScholpkopfSmola02}
Bernhard Sch\"olkopf and Alexander~J. Smola.
\newblock \emph{Learning with Kernels: Support Vector Machines, Regularization,
  Optimization, and Beyond}.
\newblock The MIT Press, 2002.

\bibitem[Shawe-Taylor and Cristianini(2004)]{Shawe-TaylorCristianini04}
John Shawe-Taylor and Nello Cristianini.
\newblock \emph{Kernel Methods for Pattern Analysis}.
\newblock Cambridge University Press, 2004.

\bibitem[Song et~al.(2012)Song, Smola, Gretton, Bedo, and Borgwardt]{SongAl12}
Le~Song, Alex Smola, Arthur Gretton, Justin Bedo, and Karsten Borgwardt.
\newblock Feature selection via dependence maximization.
\newblock \emph{Journal of Machine Learning Research}, 13:\penalty0 1393--1434,
  2012.

\bibitem[Tascikaraoglu and Uzunoglu(2014)]{TascikaraogluUzunoglu14}
A.~Tascikaraoglu and M.~Uzunoglu.
\newblock A review of combined approaches for prediction of short-term wind
  speed and power.
\newblock \emph{Renewable and Sustainable Energy Reviews}, 34:\penalty0
  243--254, 2014.

\bibitem[Tibshirani(1996)]{Tibshirani96}
Robert Tibshirani.
\newblock Regression shrinkage and selection via the lasso.
\newblock \emph{Journal of the Royal Statistical Society. Series B
  (Methodological)}, 58\penalty0 (1):\penalty0 267--288, 1996.

\bibitem[Williams and Seeger(2001)]{WilliamsSeeger01}
Christopher Williams and Matthias Seeger.
\newblock Using the nystr\"{o}m method to speed up kernel machines.
\newblock In \emph{Advances in Neural Information Processing Systems (NIPS)},
  volume~13, 2001.

\bibitem[Wilson and Vall\'ee(2002)]{WilsonVallee2002}
L.J. Wilson and M.~Vall\'ee.
\newblock {The Canadian Updateable Model Output Statistics (UMOS) systemm:
  Design and Development tests}.
\newblock \emph{Wea. Forecast.}, 17:\penalty0 206--222, 2002.

\bibitem[Yang et~al.(2012)Yang, Li, Mahdavi, Jin, and Zhou]{TianbaoAl12nystrom}
Tianbao Yang, Yu-feng Li, Mehrdad Mahdavi, Rong Jin, and Zhi-Hua Zhou.
\newblock Nystr\"{o}m method vs random fourier features: A theoretical and
  empirical comparison.
\newblock In F.~Pereira, C.J. Burges, L.~Bottou, and K.Q. Weinberger, editors,
  \emph{Advances in Neural Information Processing Systems (NIPS)}, volume~25.
  Curran Associates, Inc., 2012.

\bibitem[Zamo et~al.(2016)Zamo, Bel, Mestre, and Stein]{ZBMS16}
M.~Zamo, L.~Bel, O.~Mestre, and J.~Stein.
\newblock {Improved gridded wind speed forecasts by statistical postprocessing
  of numerical models with block regression}.
\newblock \emph{Weather and Forecasting}, 31:\penalty0 1929--1945, 2016.

\bibitem[Zhang et~al.(2018)Zhang, Filippi, Gretton, and Sejdinovic]{ZhangAl18}
Qinyi Zhang, Sarah Filippi, Arthur Gretton, and Dino Sejdinovic.
\newblock Large-scale kernel methods for independence testing.
\newblock \emph{Statistics and Computing}, 28, 2018.

\end{thebibliography}
%
\newpage
\appendix
\noindent{\LARGE \bf Experimental details}
\vspace{0.5cm}
\\
This short appendix is dedicated to the full description of the parameters that we use in the experiments.

\section{Importance of variables and their evolution through time}

\subsection{LASSO}
The main parameter of the LASSO is the regularization intensity $\lambda$. For each data split, we select it based on the NRMSE achieved on the validation set. We consider values in a geometric grid of size $30$ ranging from $10^{-5}$ to $1$. 

\subsection{BAHSIC}
We use a Gaussian kernel for both for the input kernel and the output one:
$$k_{\gamma}(\mathbf z, \mathbf z') := \exp \left ( - \gamma (\| \mathbf z - \mathbf z' \|_2^2 \right ).$$
We follow \citep{SongAl12} in the choice of the parameter $\gamma$. We standardized both our input and output data so we can apply their heuristic: set this parameter to $\frac 1 {2 d}$ where $d$ is the dimension of the inputs of the kernel. Then for the input kernel we have $d=q$ and for the output one $d=m$.

For the Nyström approximation, we use fewer points than for the KRR since as highlighted in \citep{ZhangAl18}, for detection of dependency, a fewer number of anchor points are generally sufficient. We then use $100$ points for both the input and output approximation.

\section{Wind speed and wind power forecasting}

As a first general note, since we standardized all the variables, we consider the same parameter ranges for prediction of wind speed and wind power. 
\vspace{0.17cm}
\\
\noindent{\bf LASSO}
The fitted models used for interpretation in the variable selection section are the same that we use here (so the considered parameters are the same). 
\vspace{0.17cm}
\\
\noindent{\bf OLS f-stagewise}
We selected on the validation set the number of included variables. We consider the following number of variables:
$\{5, 6, 7, 8, 9, 10, 11, 12, 13, 14, 15, 20 \}$.
\vspace{0.17cm}
\\
\noindent{\bf Nystr\"om KRR}
We select both the input Gaussian kernel's $\gamma$ parameter and the regularization parameter $\lambda$. We consider the following values: 
\begin{itemize}
\item $\gamma$ in a geometric space of length $30$ ranging from $10^{-6}$ to $10^{-3}$. 
\item $\lambda$ in a geometric space of length $30$ ranging from $10^{-4}$ to $5$. 
\end{itemize}
For the Nystr\"om approximation, we use $300$ sampled points. 
\vspace{0.17cm}
\\
\noindent{\bf Xg-Boost}
For Xg-Boost, we validate the trees' maximum depth considering  values in $\{ 3, 4, 5, 6 \}$ as well as the minimum loss reduction parameter for values in a geometric space of size $50$ ranging from $10^{-7}$ to $50$.
\vspace{0.17cm}
\\
\noindent{\bf Feedforward NN}
We consider a NN with $3$ hidden layers and validate the number of neurons per layer choosing among the possible values $\{(35, 20, 5), (50, 25, 10), (50, 35, 20), (75, 50, 25) \}$.

\end{document}